# Trends in Vehicle Re-identification Past, Present, and Future: A Comprehensive Review

Zakria[1*], Jianhua Deng[1], Muhammad Saddam Khokhar[2], Muhammad Umar Aftab[3], Jingye Cai[1*], Rajesh Kumar [4] and Jay Kumar [4]

[1] School of Information and Software Engineering, University of Electronic Science and Technology of China, Chengdu, Sichuan 610054, China

[2] School of Computer Science and Communication Engineering, Jiangsu University, Zhenjiang 212013, China

[3] Department of Computer Science, National University of Computer and Emerging Sciences, Chiniot-Faisalabad Campus, Chiniot 35400, Pakistan

[4] School of Computer Science and Engineering, University of Electronic Science and Technology of China, Chengdu, Sichuan 610054, China

[a]Corresponding author; Jingye Cai. Email: jycai@uestc.edu.cn, Zakria. Email: zakria.uestc@hotmail.com

## Abstract

Vehicle Re-identification (re-id) over surveillance camera network with non-overlapping field of view is an exciting and challenging task in intelligent transportation systems (ITS). Due to its versatile applicability in metropolitan cities, it gained significant attention. Vehicle re-id matches targeted vehicle over non-overlapping views in multiple camera network. However, it becomes more difficult due to inter-class similarity, intra-class variability, viewpoint changes, and spatio-temporal uncertainty. In order to draw a detailed picture of vehicle re-id research, this paper gives a comprehensive description of the various vehicle re-id technologies, applicability, datasets, and a brief comparison of different methodologies. Our paper specifically focuses on vision-based vehicle re-id approaches, including vehicle appearance, license plate, and spatio-temporal characteristics. In addition, we explore the main challenges as well as a variety of applications in different domains. Lastly, a detailed comparison of current state-of-the-art methods performances over VeRi-776 and VehicleID datasets is summarized with future directions. We aim to facilitate future research by reviewing the work being done on vehicle re-id till to date.

## 1. Introduction

Due to growing global population, commercial activities have been extensively increasing, which leads everyone to access road transportation as a source of mobility. Due to easy accessibility of road transportation system, traffic on roads is massively increasing that not only creates the problem of high traffic congestion but also a drastic increase in carbon dioxide emissions. Along with these issues, road accident risks and the overall transportation complexity increases as well. Therefore, a smooth transportation source and medium is always required for growing commercial activities. Furthermore, traffic management authorities are facing hectic challenges to maintain an undisturbed transportation system. Their task includes tracking the suspicious vehicle, handling traffic jam, and to check whether the vehicle is registered or not. Maintaining undisturbed transportation becomes harder when a large number of vehicles are on the roads.

## 1.1 Intelligent Transportation System

Transport is essential for the daily routine functioning of the economy and the society. Over the past few decades there is huge development, deployment, and growth in the transport system and have notable effect of development in society and daily life. Therefore, transportation should be redefined as ITS. Currently, not only mechanical and engineering fields are doing research and development for better transportation facility, but computer science related concepts are also playing major role for instance, artificial intelligence (AI), communication, machine learning (ML), internet and so many other emerging technologies.

Due to traffic problems in China, the average speed of vehicle has been decreased to 20 km/h, even in some areas between 7 and 8 km/h [1][2]. Such low speed of vehicles for a long time on roads is a threat for the natural environment of the world like exhaust emissions that deteriorate air quality. In order to deal traffic problems and alleviate the pressure of vehicles on roads, the governments are investing too much on research and ITS development. ITS based infrastructure strengthens the relationship between people, vehicles, and road networks.

ITS have the capability to enhance the performance of current transportation system and make it efficient, safe, comfortable as well as reduces harmful environmental consequences. ITS based real-time applications include electronic payment systems, traffic management systems, emergency vehicle pre-emption management system, advanced vehicle control systems, weather precautionary measures management system, and commercial vehicle operations. Applications of ITS now regularly deployed, such as closed-circuit television surveillance, automatic car parking, electronic toll collection, border control, and in-car navigation equipment. Therefore, an ITS is needed to analyze the recorded video, control, maintain and communicate to ground transport and improve mobility and manage problems efficiently. Furthermore, Figure 1 demonstrates the ITS based environment.

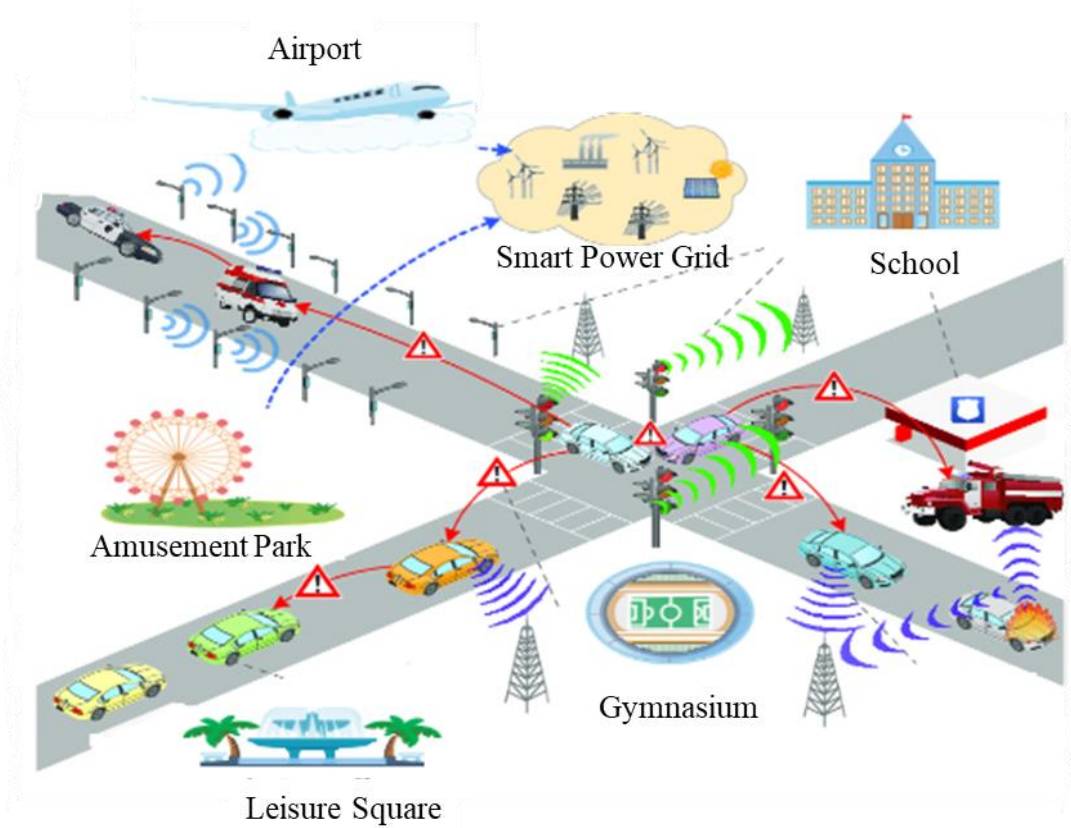

Figure 1. Depicts smart city and intelligent transportation system.

## 1.2 Video Surveillance

In metropolitan cities, cameras are widely adopted in numerous areas to monitor activities [3]; but most of the current video surveillance systems provides the facilities like capture, storage and distribute video, while leaves unwanted event detection task totally on human operators. Human operator-based monitoring of the surveillance system is not as efficient and a very labour-intensive task, as shown in Figure 2. It requires full visual attention by watching the video in control room and it is very difficult for single person as everyday tasks. Specifically, the ability to focus and react to occasionally occurring activities that require full attention. Furthermore, millions of hours of video data generated by multiple cameras over surveillance network require large number of operators for the task. It's almost infeasible, inefficient and costly to obtain real-time prevention.

Due to digital cameras and the advent of powerful computing resources, automatic video analysis become possible and more and more common in video surveillance applications [4], thus reduces the labor cost. Practically, the objective of automatic video analysis for safety, security, and surveillance is to detect automatically unwanted events or situations that need security attention. Automated video analysis not only process the data faster but also significantly improve the ability to preempt incidents on time. Augmenting security staff with automatic processing increases their efficiency and effectiveness. For the posterior mode, searching a specific vehicle in hundreds of hours of camera recorded video footage needs large number of officers to do this task and takes a lot of time. Automated content-based video retrieval reproducing and assisting human analysis on recorded videos largely enhances forensic capabilities. Furthermore, the surveillance systems application's main goal is to develop intelligent systems that automate the human decision-making mechanism.

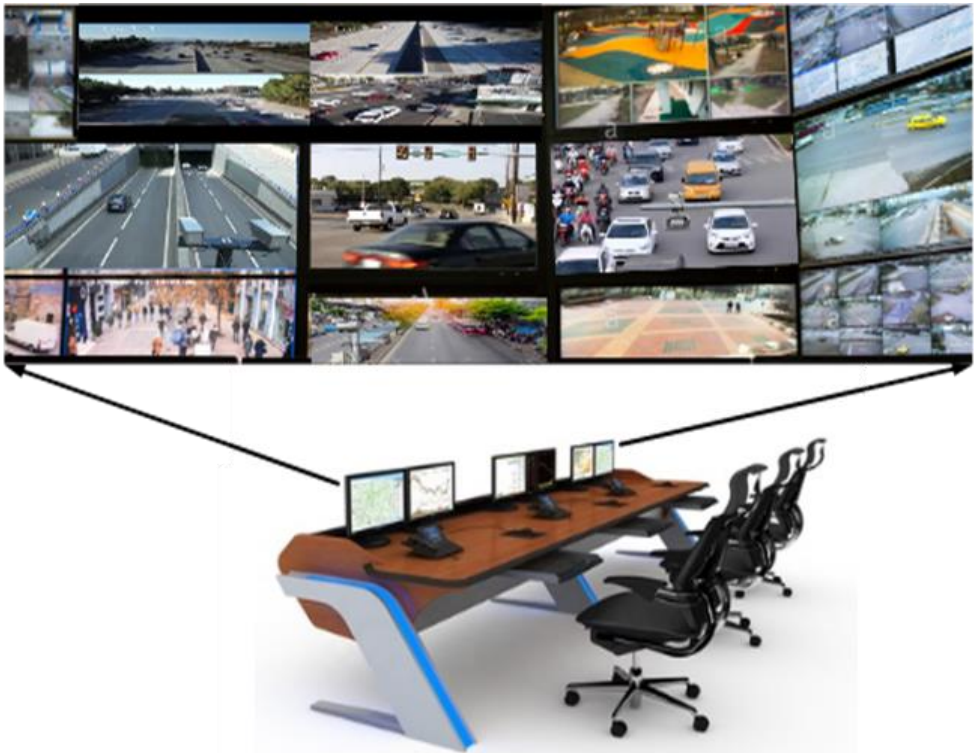

Figure 2. Shows view of manually traffic monitoring at control room.

An important task to maintain a smooth transport system is to re-identify the specific vehicle that appeared in different cameras over the surveillance network. The vehicle re-id module in ITS should recognize same vehicle that appears in surveillance cameras installed in different geographical locations. Specifically, vehicle re-id can be treated as a fine-grained recognition problem [5][6] that identifies the subordinate type of input class. However, the vehicle re-id problem's granularity is much finer since the system should search specific targeted vehicle instead of the same vehicle model and type. Moreover, recently vehicle re-id gained more attention in research community because of various significant real world applications. It is a difficult task to analyze the surveillance environment for effective vehicle identification. An example of practical environment can be seen in Figure 3, where surveillance cameras can be observed over roads and public places.

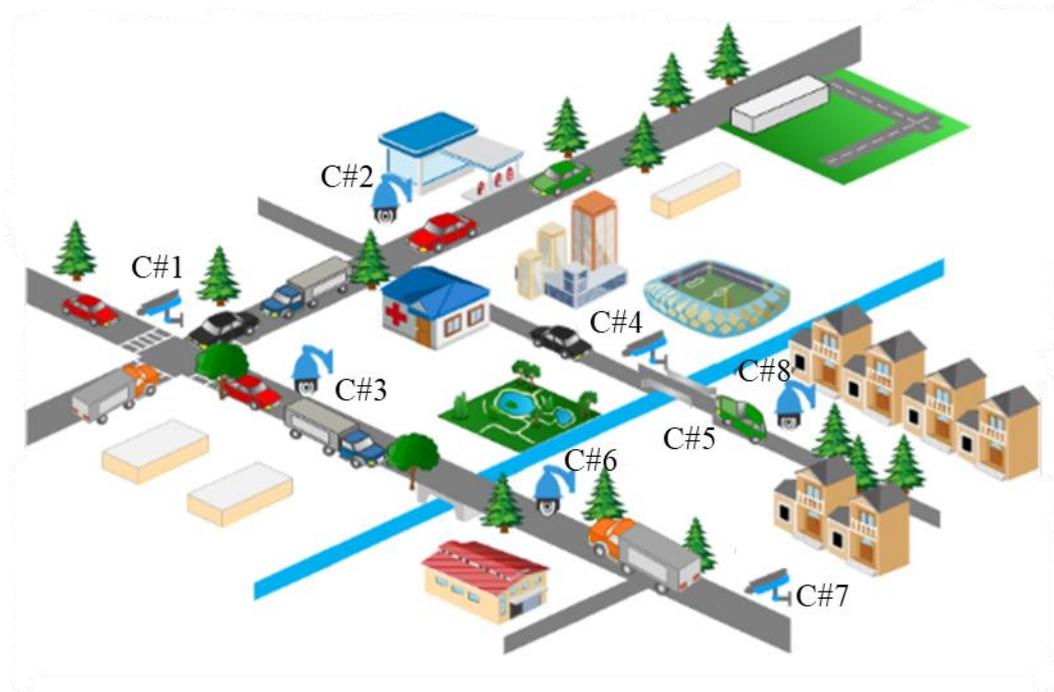

Figure 3. Illustrates the practical scenario of surveillance camera network.

## 1.3 Re-identification

In a surveillance camera without overlapping vision, re-id is defined as a task to identify objects' captured images taken from different camera networks. It is used to know whether the object image captured by multiple surveillance cameras matches the same object or a different image of the object. Object re-id technology has a significant role in multi-object tracking, intelligent monitoring, and other fields. Recently, re-id gained extensive attention in the computer vision research community. The main application fields of an object re-id are vehicle re-id and person re-id.

Formally, re-id can be defined as a matching task. A targeted image (Query) is matched against a gallery set image (representing the previously captured images in the surveillance camera network). Thus, the query of re-identifying targeted image can be defined by its descriptor P, and it is formulated as:

$$T = arg_{T_i} \min D\,(T_i, Q), T_i \in \mathcal{T} \tag{1}$$

where $\mathcal{T} = \{T1, \ldots\ldots, TN\}$ is a gallery set of N image descriptors, and D(,) represents the distance metric. Therefore, to solve above the re-id problem, it is important first to answer how we can represent targeted object using a descriptor to robust performance. Furthermore, rest of the paper investigates this topic.

*Person Re-identification:* Person re-id is a task to determine whether the person captured in one camera is already appeared over a multiple surveillance camera network or not. Person re-id is indispensable in establishing constant labels across different cameras or even on the same camera to re-establish lost tracks. Person re-id is an extremely challenging job due to the different camera appearance response of the same person. Sometimes, the camera presents large differences compared to a different person. The change in appearance is usually because of significant variation in illumination, viewpoint, occlusion, color, resolution, so on across different camera views. After appearance other feature for person R-Id is biometric cues, such as gait and face. Still, these cues are usually infeasible in real-world camera surveillance systems due to different cameras resolution, a different perceived posture of the person like sitting, standing and sleeping that can easily fool the person re-id model and make it unusable.

Mostly person re-id approaches have concentrated on appearance-based models [7][8] and learning-based methods [9][10]. In appearance-based models, the task is to extract robust feature vector to encounter appearance challenges such as occlusion and view changes in camera network. Whereas, learning-based model system is trained to calculate the distance between objects' features. If the distance is low, it means that the object is same, otherwise different. Recently, neural network outperformed on feature learning and matrix learning for person re-id [9][10][11][12].

Its practical applications ranges from tracking specific person across multiple cameras network to search him in a large gallery set, from customer analysis in a retail outlet to identifying specific person from group photo; forensic search, multi-camera behavior analysis and analyzing customer shopping trends by observing their touching, surveying, and trying products in stores under different surveillance cameras. Another example is that geriatric health care analysis explores the elder people's long-term behavior to assist doctors in making more accurate diagnoses. Moreover, Figure 4 shows basic steps of person re-id system.

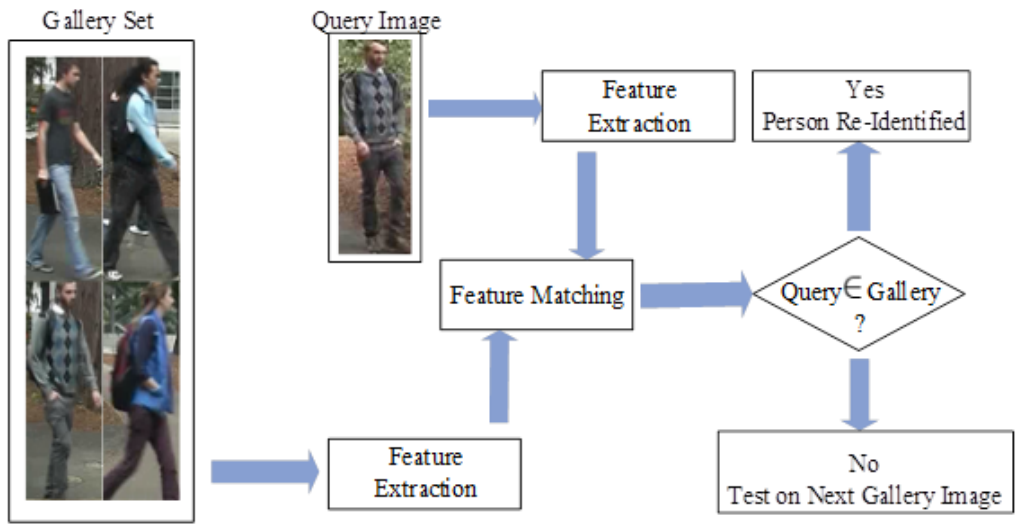

Figure 4. Represents the standard structure of person re-id system.

***Vehicle Re-identification:*** Similar to person re-id, vehicle re-id is also a demanding task in camera surveillance. Aim of vehicle re-id is to match vehicle images with already captured vehicle images over the camera network [13][14][15]. However, due to surveillance cameras on the roads for smart cities and traffic management, the demand to perform vehicle search from the gallery set is increased. Vehicle re-id is similar to several other applications, such as person re-id [16], behavior analysis [17], cross-camera tracking [18], vehicle classification [19], object retrieval [20], object recognition [21][22], and so on.

To understand designing the vehicle re-id system, we analyze how a person re-identifies the vehicle. A person re-identifies vehicle by keeping in mind some characteristics like unique feature, color, size etc., our brain and eyes are learned to detect and identify different objects, as shown in Figure 5 and how system identify vehicle is shown in Figure 6.

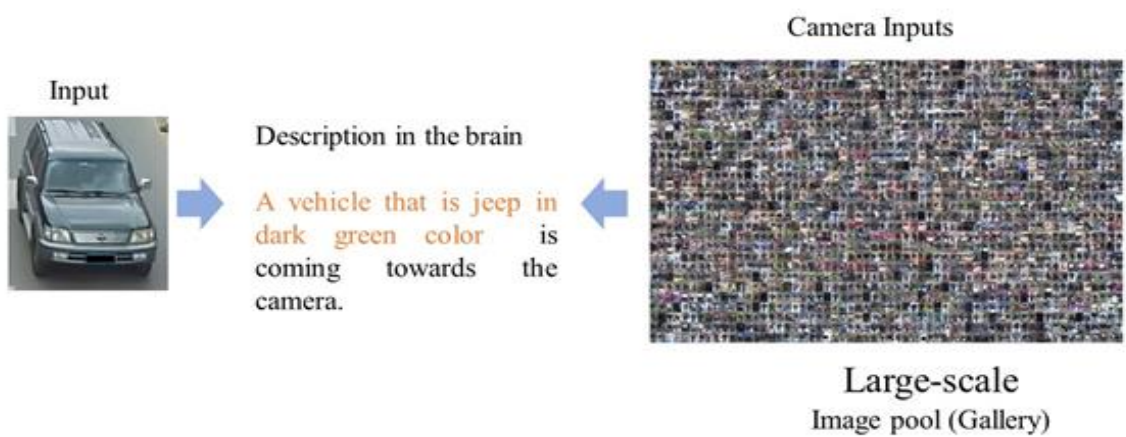

Figure 5. Shows how human re-identify vehicle?

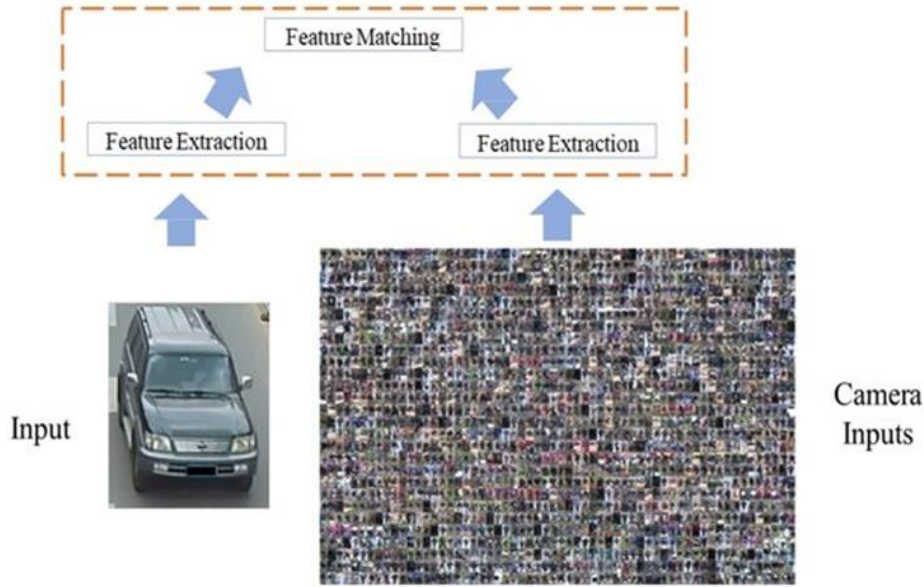

Figure 6. Illustrates how machine re-identify vehicle?

***Person Re-identification vs Vehicle Re-identification:*** Both person and vehicle re-id tasks are almost similar, but comparatively vehicle re-id is more complicated and challenging than person re-id. Furthermore, directly deploying person re-id approaches for vehicle re-id may not perform well. Vehicle and person images captured by multiple camera network in different viewpoint can be seen in Figure 7. It can be observed from that the difference (based on visual features) between two viewpoints of a vehicle is much higher than the two viewpoints of a person. It is worth noticing that the vehicle's visual patterns are non-overlapping for front, rear, and side viewpoints. In contrast, the visual appearance of a person is shared over different viewpoints. In fact, the color and texture of human body clothes normally do not change as much by changing the viewpoint, because of upright body posture. Initially proposed approaches for person re-id did not examined different viewpoints of a person while processing the data. Moreover, such person re-id approaches splitted the human body vertically for feature extraction [23][24]. However, horizontal changes are not largely considered. Therefore, conventional methods for person re-id are not appropriate for vehicle re-id. Furthermore, some methods adopt subspace learning with distance matrix [25][26] and neural networks [7][10] in brute-force manner where the distance between same person images is low, and the difference between different person images is high. In this case, we cannot apply brute-force for vehicle re-id because images of same vehicle captured from different viewpoints look like a different vehicle. On the other hand, the images of two different vehicles captured from same view angle look more alike.

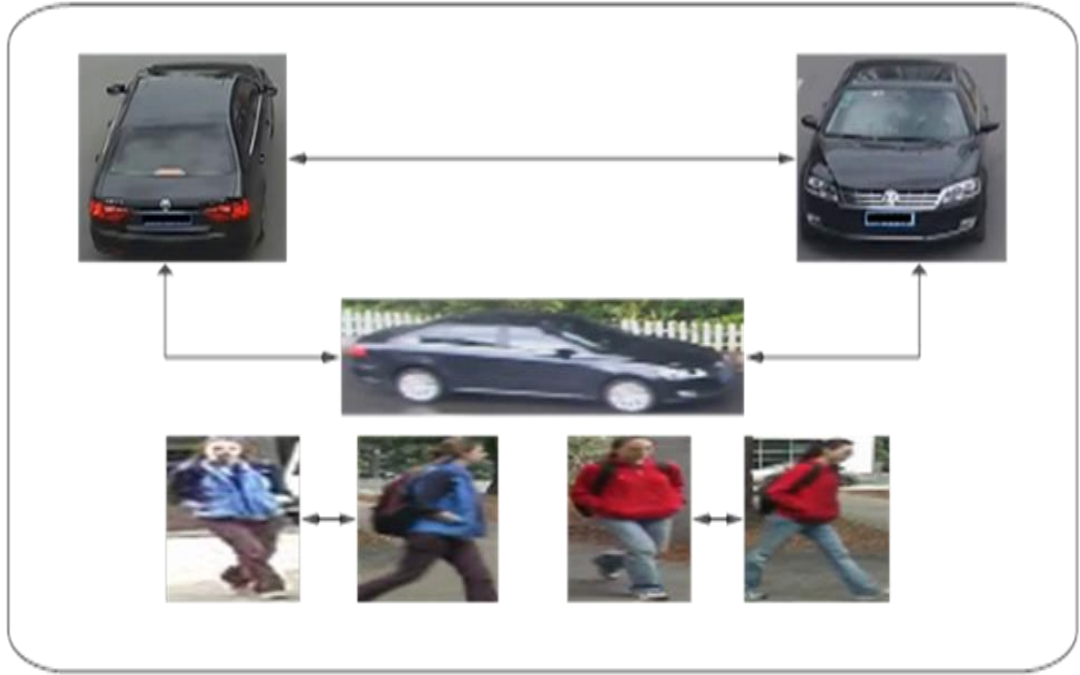

Figure 7. Show different viewpoints effect on a vehicle and a person.

## 1.4 Vehicle Re-identification Practical Application

There are many significant real-world applications where vehicle re-id system can be utilized and satisfies the great needs of our practical life. However, some major applications are briefly discussed as follows:

i.   Suspicious vehicle search: Most of the time terrorists use vehicle for their criminal activities and soon leave that spot on vehicles. It is very difficult to fast search suspicious vehicle manually from surveillance camera.

ii.  Cross camera vehicle tracking: In vehicle race sports, some of the viewers on television wish to watch specific vehicle. With vehicle re-id system broadcaster can only focus on that specific vehicle when it comes in the field of view of surveillance camera network.

iii. Automatic toll collection: Vehicle re-id system can be used at toll gates to identify vehicle type like small medium and large and charge the toll rate accordingly. Automatic toll collection reduces delay and improves the toll collection performance by saving travelers time and fuel consumption.

iv.  Road access restriction management: In big cities, heavy vehicles like trucks are not permitted in the daytime, or some of the vehicles with specific license plate number are permitted on specific days to avoid congestion in city or officially authorized vehicles can enter in city.

v.   Parking lot access: vehicle re-id system can be deployed at the gate of parking lot of different places like head offices, and residential societies. So only authorized vehicles are allowed to park.

vi.   Traffic behavior analysis: Vehicle re-id can be used to examine the traffic pressure on different roads at different times, such as peak hours calculation or particular vehicle type behavior.

vii.   Vehicle counting: System can be useful to count a certain type of vehicle.

viii.   Speed restriction management system: Vehicle re-id system can be utilized to calculate the vehicle's average speed when it is crossing from two subsequent surveillance camera positions.

ix.   Travel time estimation: Travel time information is important for a person who is traveling on road, it can be calculated when a vehicle is passing in between consecutive surveillance cameras.

x.   Traffic congestion estimation: By knowing the number of vehicles flow from one point to another point within a specific time period using vehicle re-id system, we can estimate traffic congestion at the common spot from where all vehicles may cross.

xi.   Delay estimation: Specific commercial vehicle delay can be estimated after predicting traffic congestion on the rout that vehicle follows.

xii.   Highway data collection: Highway data can be collected through surveillance cameras that are installed on roadsides and that data can be used for any purposes after processing and analyzing at the traffic control center.

xiii.   Traffic management systems (TMS): Vehicle re-id is an integral part of TMS, it helps to increase transportation performance, for instance, safe movement, flow, and economic productivity. TMS gathers the real-time data from the surveillance cameras network and streams into the Transportation Management Center (TMC) for data processing and analyzing.

xiv.   Weather precautionary measures: When specific vehicle is identified that may be affected by weather, then traffic management systems notify that vehicle about weather conditions like wind velocity, severe weather etc.

xv.   Emergency vehicle pre-emption: If any suspicious vehicle is identified at any event or road then vehicle pre-emption system passes messages towards lifesaving agencies such as security, firefighters, ambulance, traffic police, etc. to reach in time and stabilize the scene. With this system, we can maximize safety and minimize response time.

xvi.     Access control: Vehicle re-id system can be implemented for providing safety and security, logging and event management. With the implementation of the system only authorized members can get an automatic door opening facility, which helps guards on duty.

xvii.    Border control: Vehicle re-id system can be adopted at different check posts to minimize illegal vehicle border crossing. Vehicle re-id system can provide vehicle and owner's information as it approaches security officer after identifying the vehicle. Commonly these illegal vehicles are involved in cargo smuggling.

xviii.   Traffic signal light: When the traffic light is red and any vehicle crosses stop line, the vehicle re-id system can be implemented to identify that vehicle for fine.

xix.     Vehicle retrieval: In this case, re-id is associated with a recognition task. The specific query with a target vehicle is provided, and all the related vehicles are searched in the database. The re-id task is thus employed for image retrieval and usually provides ranked lists, similarly related items, and so on.

However, due to the vast range of practical applications that employ vehicle re-id system and to limit the scope of the paper, this review article mainly focuses only on vision-based methods. Moreover, it is very hard to cover all technologies for vehicle re-id in one survey paper but despite of that we have summarized the strengths and weaknesses of all technologies in Table 1. Therefore, this review article focuses on the use of vision-based approaches including, Appearance, license plate, contextual information etc. In last few years, there has been lack of comprehensive study of the overall problem and different solutions. This paper fills the gap by providing a detailed review covering main challenges, different approaches, and applicability. In addition, it provides the analysis and comparison of existing vehicle re-id methodologies. Aiming to facilitate other researchers, this review also provides the required information about the publicly available datasets and discusses several important research directions with under-investigated open issues to narrow the gap between the closed-world and open-world applications, taking a step towards real-world re-id system design.

Two ways for writing surveys can be found in the object re-id literature; first way gives a deep insight into methodologies, whereas second way covers the overall perspective related to the problem [27][28]. This survey includes both methodologies and overall perspective of vehicle re-id literature. We also review recent development of vision based vehicle re-id along with other technologies. In addition, this survey draws a timeline to introduce important milestones for vehicle re-id, which can be seen Figure 8.

The paper is organized in the following way. Section 2, 3, 4,5 provides an overview of recent state-of-the-art proposed methodologies in various technologies. Section 6 presents a publically available benchmark dataset that cover various real-world surveillance scenarios. Section 7 discusses the challenging problems in vehicle re-id. Section. 8 sheds light on the evaluation measures for vehicle re-id. Section 9 analyzes and compares the experimental results of various approaches. Meanwhile, the last section concludes and discusses future work.

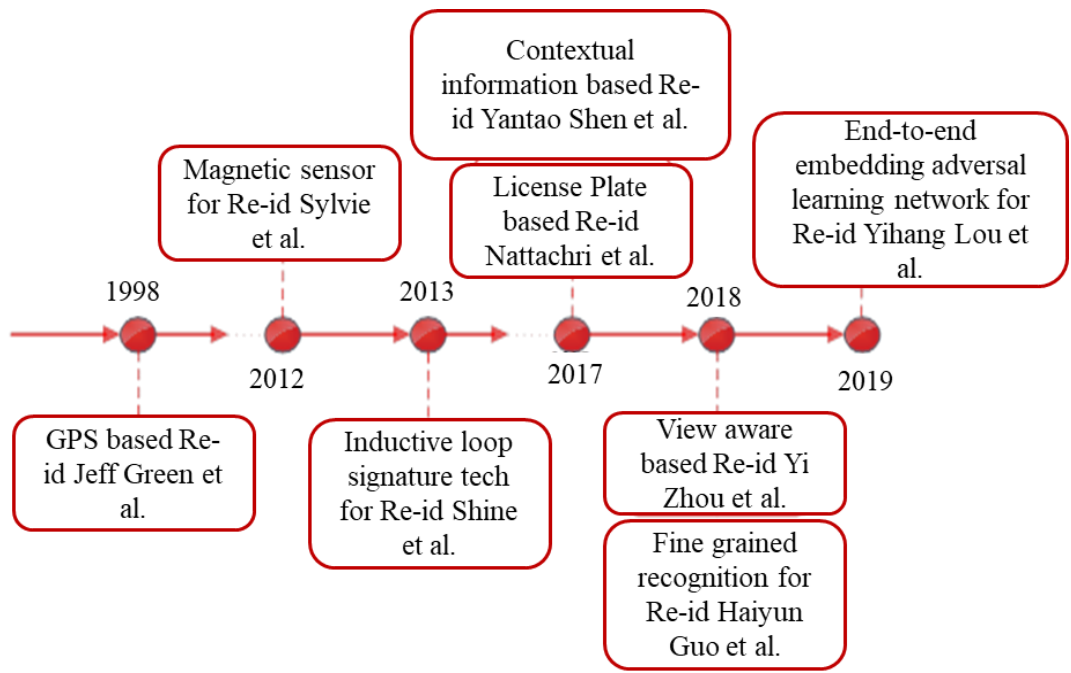

Figure 8. Milestones existing re-id approaches in the Vehicle re-id history.

The main contributions of this review paper is summarizes as follows:

i. To the best of our knowledge, this is the first comprehensive review paper that cover computer vision based methods for vehicle re-id tasks, with different technological background of approaches for completeness such as, global positioning systems (GPS), inductive loop and magnetic sensors.

ii. Discusses various real-world applications of vehicle re-id in different domains including the intelligent transportation system.

iii. Comprehensive comparisons of existing methods on several state-of-the-art publicly available vehicle re-id datasets are provided (e.g., in Tables 1,2, 3, 4, Figures 34, 35), with brief summaries and insightful discussions being presented.

iv. Discusses the challenges in detail for designing an efficient vehicle re-id system and illustrates the recent trends and future directions.

## 2. Methods Used for Vehicle Re-identification

Traditionally different traffic sensors are adopted to know the vehicle presence, volume, occupancy, and speed data. Nowadays, new sensor-based technology is adopted to get more information like origin-destination estimation, travel time and other travel information applications. Based on different technologies vehicle re-id approaches can be divided into six categories, as depicted in Figure 9.

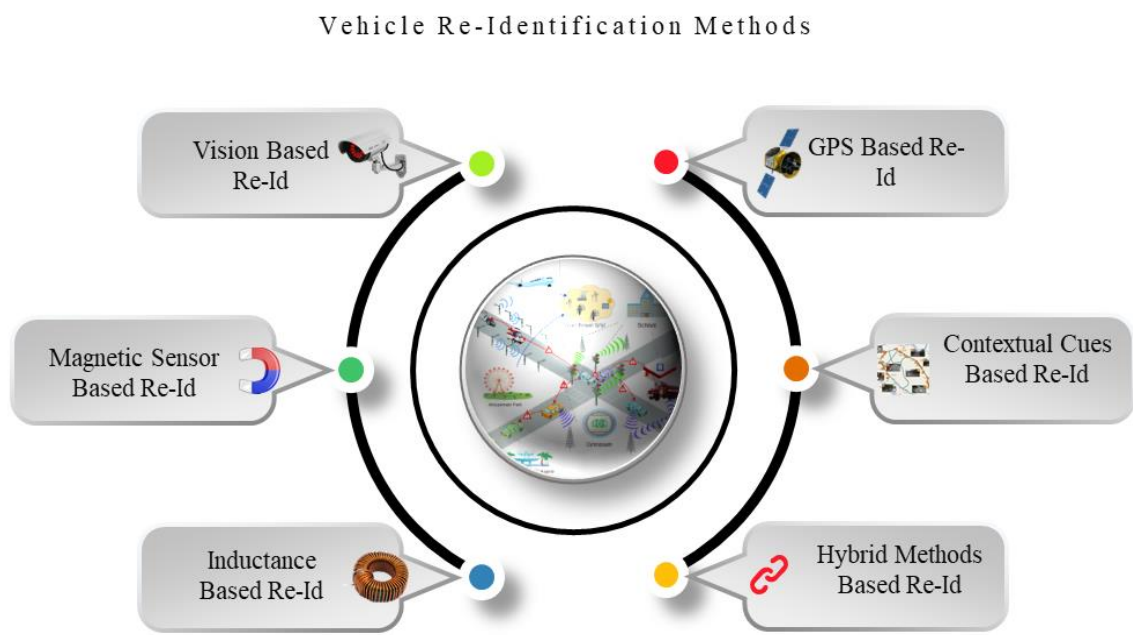

Figure 9. Shows vehicle re-id methods.

## 2.1 Magnetic Sensor-based Vehicle Re-identification

An electromagnetic field is used to detect the vehicle, when it crosses and it is used to provide occupancies, counts, and vehicle speed. However, vehicles are made up of metal. It disrupts the magnetic field, so magnetic signature regenerated by one vehicle is different from the other vehicle [25]. This approach helps in re-identifying a specific vehicle. Moreover, for ITS the Berkeley's company sells magnetic sensors with the name "Sensys Network" [29]. A straight-line re-id rat is 50%, and the approach reduces the magnetic signature peak value sequence for calculating the signature distance to prevent vehicle speed dependency [30]. For real-time vehicle re-id processing unit is associated to thousands of magnetic sensor nodes and a large number of magnetic sensors that generate massive data streams, and to deal with real-time data stream mining, high-performance FPGAs and low-performance microcontroller are used [31][32]. Sylvie Charbonnier et al. [33] studied various approaches for vehicle re-id by adopting vehicle tridimensional magnetic signature measured with sensor, when car passes sensor and changes in the magnetic field were induced and measured in three different directions like X, Y, Z. Rene O. Sanchez et al. [34] investigated vehicle re-id approaches by using wireless magnetic sensors and compares vehicle magnetic signatures to overcome the limitations of system while vehicle is stopped or moving slow at detection station.

## 2.2 Inductive Loop-based Vehicle Re-identification

Vehicle can be re-identified using inductive loops embedded in the road surface for the detection of vehicle. From those loops, a fingerprint is captured for every car passing by. The travel time can be determined when those fingerprints or certain aspects of them coming from different locations are compared with each other. Jeng and Chu [35] designed a real-time inductive loop signature-based vehicle re-id method named RTREID-2M. Inductive signature is used for vehicle re-id and much efforts have been done to utilize inductive loop signature technology. Inductive signature based vehicle re-id algorithms identify specific vehicle at downstream detection station by matching the inductive signature at upstream detection station, considering that vehicle have same signature by crossing different loop detection stations [36]. Vehicle re-id researchers have proposed several algorithms like optimization [36], piecewise slope rate (PSR) matching [37], lexicographic and blind deconvolution [38], all these proposed approaches are for raw signature processing, signature feature extraction, and vehicle matching. R.J. Blokpoel [39] proposed an algorithm with different sizes of a single loop. Validation tests depict re-id rates up to 100%, when loops are identical to the similar type and 88% when compare between different types.

## 2.3 Global Positioning Systems-based Vehicle Re-identification

Global Positioning Systems (GPS) technology is an essential and valuable tool for ITS and traffic surveillance, because it provides positioning data for every single vehicle [40][41]. There are still some limitations in vehicle re-id using GPS like varying accuracy, minimal fleet penetration, and signal loss because of tunnels, trees, tall buildings, etc. GPS is adopted with vehicles to locate and get travel information along with longitude and latitude information and timestamp. GPS is special form of mobile sensing technology that enables the devices like GPS logger, GPS cellular phones, and smartphones moves with vehicles to get speed information and location continuously. However, different types of vehicles have different behaviors such as deceleration rates, acceleration, and speed variation. This encourages the author to adopt GPS technology for vehicle classification and re-id [42]. Figure 10 shows the example of a vehicle equipped with GPS.

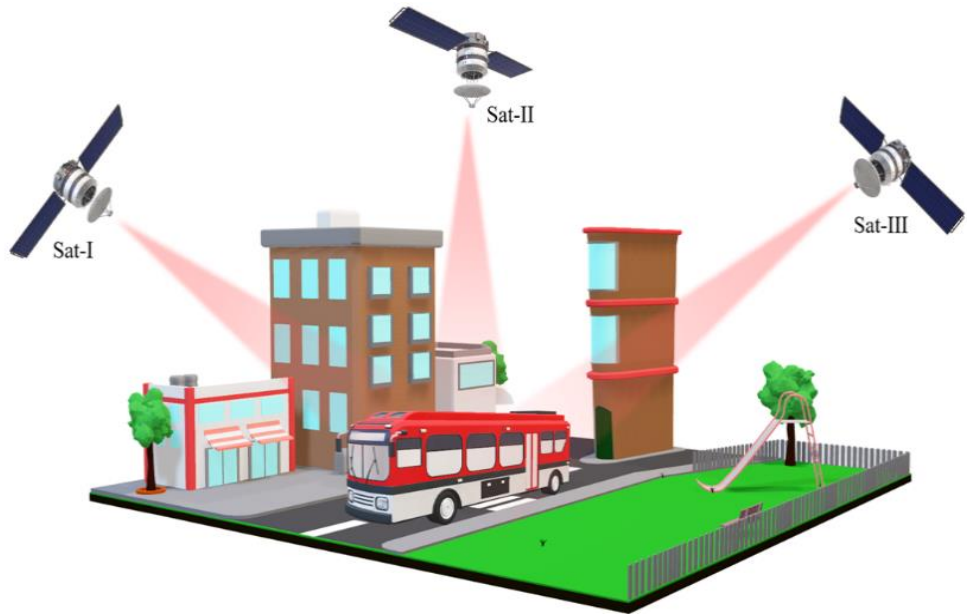

Figure 10. A vehicle equipped with GPS.

## 2.4 Vision-based Vehicle Re-identification

In computer vision, the aim of vehicle re-id is to identify specific vehicle that appeared over in multiple cameras network. The large surveillance camera network is deployed in different areas of public places like hospitals, parks, colleges, roads, and other areas. It is also difficult and tiresome job for security officers to track targeted or specific vehicle over multiple camera network manually. However, computer vision techniques can automatically re-id a vehicle and basic five main working steps are discussed below (shown in Figure 11).

i. Step 1: Data Collection: For real-time video analysis raw videos from surveillance cameras is one of the key component. The cameras are fixed at different locations in an unconstrained environment [43].

ii. Step 2: Bounding Box Generation: It is very difficult almost impossible when we have large scale surveillance videos to extract vehicle image. We use a bounding box and it is obtained by vehicle detection technique [44][45].

iii. Step 3: Training Data Annotation: Data annotation is a process of labeling the videos or images of dataset with metadata. It is an indispensable step for vehicle re-id model training because each surveillance camera video recording is in a different environment.

iv. Step 4: Model Training: Model training is simply the task of learning discriminative features and good values for all the weights and the bias from previous annotated vehicle videos or images of the dataset. It is a key step in vehicle re-id systems and a widely explored area in literature.

v. Step 5: Vehicle Retrieval: Vehicle retrieval is a task of matching targeted vehicle (query image) over a gallery set.

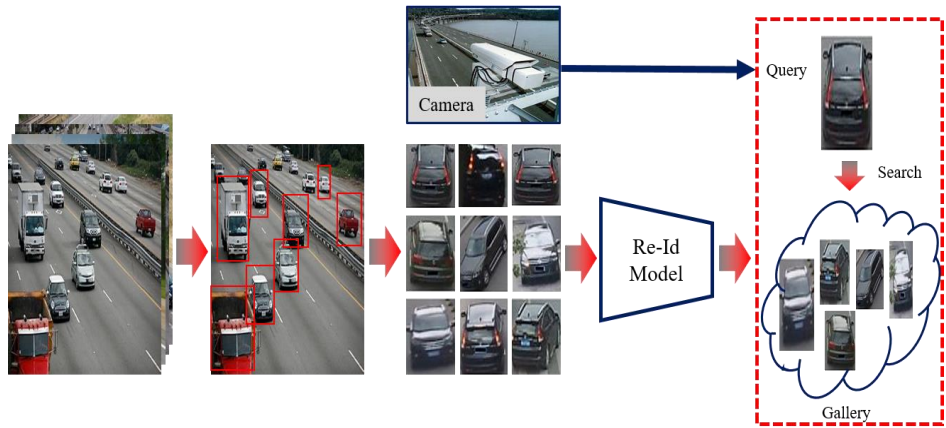

Figure 11. The flow of designing a practical vehicle re-id system, including five main steps.

Table 1 Summary of strengths and weaknesses of different vehicle re-id technologies.

| Technology | Strengths | Weaknesses |
|---|---|---|
| Surveillance camera | • Don't require the owner's cooperation.<br>• Low cost because usually cameras are installed on roadsides, so don't require additional charges to install. | • Complex and unconstrained environment along with varied road topology affects the performance.<br>• Performance degrades due to dirt, snow, occluded image, blurry image, and sunshine, etc.<br>• The vehicle is identified only when it comes in the field of view of the camera. |
| Magnetic Sensor | • Insensitive to bad weather like snow, fog, and rain.<br>• There is no privacy issue in magnetic sensors. | • Complicated installation.<br>• Embedding magnetic sensor under carriageway after drilling hole.<br>• Identified only at the detection terminal. |
| Inductive loop | • Provides different traffic parameters like speed, volume, headway, presence, and occupancy etc. | • Installation of inductive loop technology requires metallic loops under the road.<br>• The vehicle is identified in the field-of-view of detection terminal. |
| GPS | • Provides continuous vehicle information, such as space and time, to the control centre.<br>• 100% vehicle recognition rate. | • Require owner's cooperation to install hardware in vehicle.<br>• Varying accuracies, minimal fleet penetration, and signal loss because of tunnels, trees, and tall buildings. |

## 3. Vision-based State-of-the-Art Vehicle Re-identification Approaches

Vision-based methods focus on examining robust feature representations to calculate the distance between features of two-vehicle images and vehicles with the same class have a low distance otherwise high. However, vehicle features are difficult to distinguish when a captured vehicle image consists of similar colors and pose. In this section gives an overview of recent works on computer vision based methods for vehicle re-id problem. Several impressive vision-based methods have been proposed to improve vehicle re-id performance either by modifying the existing DL architectures or designing a new deep neural network (DNN). Generally speaking, eight different techniques have been employed in this research area: A) Feature representation for vehicle re-id, B) Similarity metric for vehicle re-id, C) Traditional machine learning-based vehicle re-id, D) View-aware based vehicle re-id, E) Fine-grained visual recognition based vehicle re-id, F) Generative adversarial network-based vehicle re-id, G) Attention mechanism, H) License plate-based vehicle re-id.

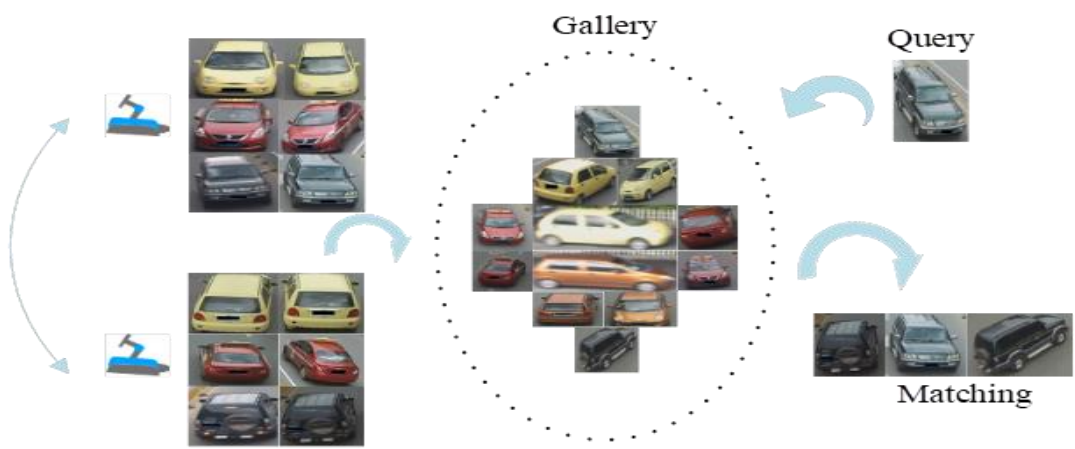

Figure 12. The vehicle re-id problem: given a Query, find the matching candidate in the gallery.

## 3.1 Feature Representation for Vehicle Re-identification

Feature representation play vital role in progress of many different computer vision tasks. In this regard, vehicle re-id features representation approaches can primarily be classified into two parts: hand-crafted and deep learning features representations. Hand-crafted feature representations BOWCN [46], and LOMO [47] initially utilized in person re-id and then applied directly on vehicle re-id task. Some well-known deep learning based feature representations such as GoogLeNet [48], VGGNet [49], AlexNet [50], and, ResNet [51] are used for vehicle re-id. The researcher also adopts these baseline models in their approaches for vehicle re-id. Such as, NuFACT [52] takes GoogLeNet [48], FACT [53] uses AlexNet [50], DRDL[54] utilizes VGGNet [49] to extract features of vehicles. Various type of loss functions are utilized to efficiently learn vehicle image discriminative feature representation to train deep learning based model vehicle re-id; such as the deep joint discriminative learning (DJDL) [55] approach uses identification, and verification and triplet loss functions improved triplet convolutional neural network [56] uses classification and-oriented and triplet loss function to extract discriminative feature representation.

## 3.2 Traditional Machine Learning-based Vehicle Re-identification

In traditional machine learning (TML), we adopt feature engineering to artificially clean and refine data. However, previously proposed approaches are grouped into for robust features extraction and learning discriminative classifiers. In TML extracted features are directly computed from image pixels and it is low level feature representation. Moreover, TML based algorithm design is expensive and difficult. Broadly, it consists of two steps feature extraction and feature classification. There are many algorithms proposed for low level feature extraction for instance speeded up robust features (SURF) [57], scale-invariant feature transform (SIFT) [58], and histogram of oriented gradient (HOG) [59]. After feature extraction different classifiers are applied, which are widely used in TML approaches such as linear regression, k-Nearest Neighbor (KNN) [60], logistic regression, support vector machine (SVM) [61], bayes classification [62], and decision tree [63]. The features extracted using SIFT are a local features of the image, which maintains the scale scaling, invariance of rotation, and brightness variation. In addition, it also maintains a particular degree of stability to affine transformation, the viewing angle change, and noise.

Moreover, one of the feature descriptor adopted for targeted object detection in image processing is HOG. The large area of image features are formed by calculating the gradient direction histograms of its local regions. However, an overlapping local contrast normalization approach is adopted to improve the performance. Zapletal and Herout [64] utilize the color histogram and the HOG features with linear regression to re-id vehicle. Chen *et al.* [65] designed a method to re-id vehicles grid-by-grid with HOG features extraction for coarse search and further improves the result by utilizing histograms of matching pairs. In [66], vehicle re-id local variance measures are applied using local binary patterns and joint descriptors.

### 3.3 Similarity Metric for Vehicle Re-identification

Performance of vehicle re-id can be improved by selecting appropriate distance matrices regardless of appearance representation. Distance metric learning approaches [67] are thoroughly studied in image retrieval and recognition tasks, in which matric space is defined in such a way that features that belong to same class are kept closer and different are at distant. In the re-id task, image features are known as appearance descriptor. In this the learned distance matric in appearance space minimizes the distance for descriptor between same vehicles and maximizes distance for descriptor of different vehicles. As in various face recognition algorithms [68][69] uses Euclidean and Cosine distance matric to measure the similarity, and FACT [53] also utilizes Euclidean and cosine distance metrics to measure similarity between the pair of vehicle for re-id. Similarly, NuFact [52] utilizes the Euclidean distance to measure the similarity between the probe and gallery set vehicle images in discriminative null space [70]. Furthermore, deep relative distance learning (DRDL) [54] studied a two-branch convolutional neural network to covert the raw vehicle images into a Euclidean space, so that distance can be used directly to measure the similarity of two individual vehicles.

Pairwise constraints are required for matrix learning and it is done in supervised fashion. During the training features of appearance descriptor are in pair and labelled as positive and negative. It is totally depending on appearance descriptor whether it belongs to the same vehicle or different vehicle. Appearance descriptors are represented as $x_1$, $x_2$,…, $x_n$, here n represents number of training instances and the dimensionality of every instance is represented by m. The aim of metrics learning is to learn distance metric and matrix $D \in R_{mxm}$ represents it; thus the distance between pair of appearance descriptors $x_i$ and $x_j$ is as follows:

$$d(x_i, x_j) = (x_i - x_j)^T D (x_i - x_j) \qquad (2)$$

d($x_i$, $x_j$) is a true metric only possible when matrix D is symmetric positive semi-definite. This issue is resolved by adopting convex programming as follows:

$$min_D \sum\nolimits_{(x_i,x_j)\in Pos}\left\|(x_i - x_j)\right\|_D^2 \, s.t. \, D \geq 0, and \sum\nolimits_{(x_i,x_j)\in Neg}\left\|(x_i - x_j)\right\|_D^2 \geq 1 \qquad (3)$$

where Pos represents the positive label in training samples, and it is the appearance descriptor of the same vehicle, whereas Neg represents the negative label in training samples and it is the appearance descriptor of a different vehicle

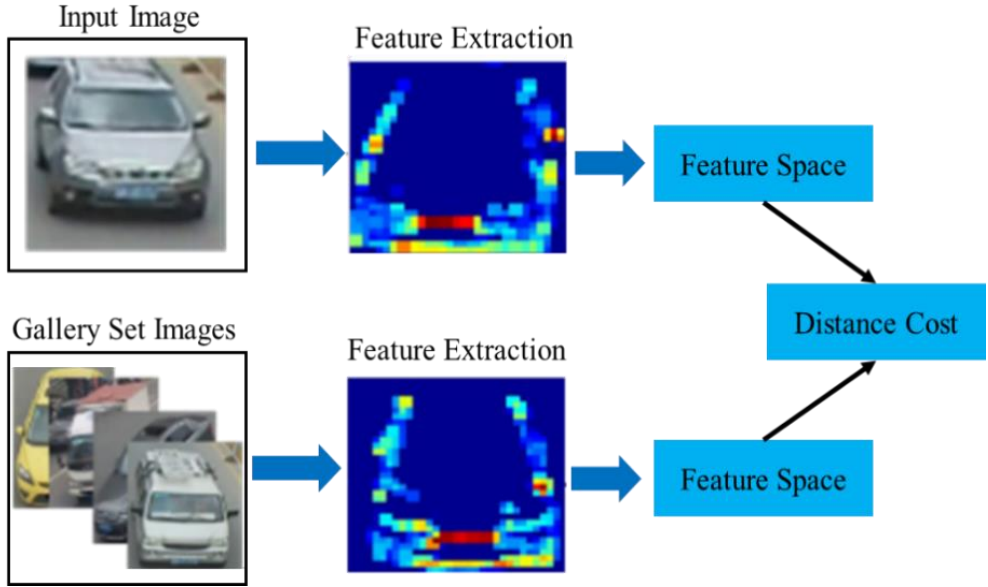

Figure 13. Vehicle re-id system based on metric-based methods.

## 3.4 Fine-grained Visual Recognition-based Vehicle Re-identification

Vehicle re-id is fine-grained recognition task, and fine-grained vehicle recognition can be divided into two parts, representation learning model and part-based model. Many approaches are proposed [71] that utilize alignment and part localization for feature extraction of main parts and then those parts are compared for vehicle re-id. Xiao et al. [72] studied weakly supervised way in fine-grained domain using reinforcement learning to get discriminative parts of vehicle. In addition, Lin et al. [73] presents a bilinear architecture to get the pair of local features in which output descriptors of two networks are merged in an invariant way. Boonsim et al. [74] presents an approach for fine-grained recognition of vehicles at night. The authors utilize shape and lights of vehicle visible in night and relative position to identify model and make of a vehicle, which are visible from the front and rear side.

In fine-grained recognition, local region features are extracted from different points such as logo, annual inspection stickers, and decorations, to make system more efficient and robust various attributes of vehicles are also incorporated like color, model, and type information. For example, in different vehicles with similar global appearance in Figure 14, all the vehicles are different in each column. The differences between each vehicle are pointed out with red circles. From Figure 14 it can also be seen that the differences between similar global appearance vehicles lie in some local regions.

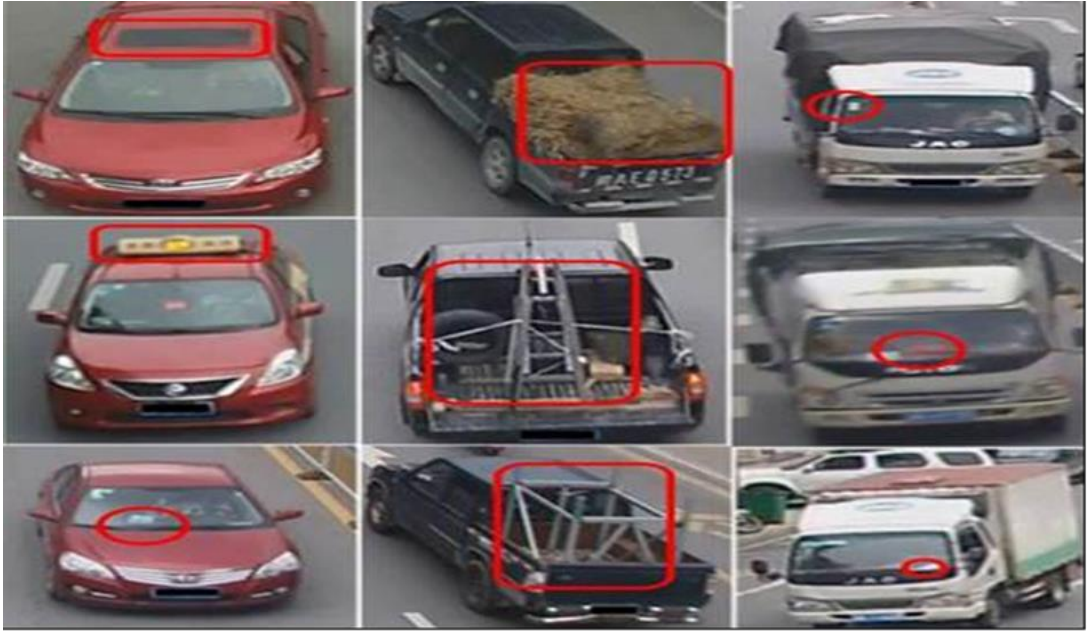

Figure 14. Shows vehicles that are same in global appearance but differentiated by local regions that are marked in red circle.

### 3.5 View-aware Based Vehicle Re-identification

Most of the above discussed deep learning features [50][49][48][55][56] are general, and these learned features end at multiple fully connected layers. Despite that, all these approaches performance is not bad. But these approaches are not designed for a specific problem related to view point variation. It is a central challenge in vehicle re-id task. Vehicle re-id is closely related to person re-id, however, intra-class variation is a major problem in person re-id in which the same person looks different by changing viewpoint. Zhao et al. [75] designed a novel approach that achieved satisfactory results and the method was based on person body parts guided for re-id. Wu et al. [76] proposed a study with pose prior that made identification efficient and robust to viewpoint. Zheng et al. [77] proposed the pose box structure that generates the pose estimation after affine transformations. It is also challenging and crucial in vehicle re-id, because image viewpoint is the same as a consequence of vehicle rigid motion. Wang et al. [78] studied the orientation invariant feature embedding to solve the issue of viewpoint variation influence on vehicle re-id system. Prokaj et al. [79] proposed a pose estimation-based approach to handle multiple viewpoint problem. Yi Zhou et al. [80] studied uncertainty in the viewpoint of vehicle re-id system and designed end to end deep learning-based architecture on Long Short Term Memory (LSTM) bi-directional loop and concatenated CNN, in this model author takes full advantage of LSTM and CNN to learn the different viewpoints of vehicle. And also, there are many more approaches are proposed to handle the view point variation issue in vehicle re-id such as adversarial bi-directional long short-term memory (LSTM) network (ABLN) [81], spatially concatenated convolutional network (SCCN) and CNN-LSTM bi-directional loop (CLBL) [80]. However, all these approaches need vehicle datasets. Every vehicle image is densely sampled camera viewpoints. Despite that, it is hard to gain in real-time camera surveillance systems. Therefore, there is still ample room for vehicle re-id by thoroughly considering viewpoint variations.

## 3.6 Generative Adversarial Network-based Vehicle Re-identification

GAN [82] is one of the hot technique in semi-supervised and unsupervised learning algorithms. It is proposed by Goodfellow by deriving backpropagation signals through a competitive process involving a pair of networks. GAN can be adopted in different applications, like style transfer, image synthesis, image super-resolution, semantic image editing, image super-resolution, classification and person/vehicle re-id. The GAN-based vehicle re-id flow is shown in Figure 16. At present, there have been many papers that adopt GAN to solve the problems of vehicle re-id. The existing datasets have low diversities and small scales, which leads to poor generalization performance on the trained models. To solve this problem. Generative Adversarial Network (GAN) in Object re-id is among the latest research trends in the deep learning approaches. GANs achieved significant performance in in many fields such as translation [83] and image generation [82][84]. Furthermore, recently GANs are also utilized for re-id problems (person re-id and vehicle re-id) [85][86]. Zheng et al. [11] proposed a method in which they used the DCGAN [84] with Gaussian noises to generate unlabeled person images before training. Wei et al. [87] studied a PTGAN to minimize the domain gap by transferring person images between different styles. Zhou et al. [88] proposed GAN based model to solve cross-view vehicle re-id problem by generating vehicle images in different viewpoints. Lou et al. [85] designed a model to generate the same and cross-view vehicle images from original images to facilitate training model. Zhou et al. [88] proposed a conditional generative network to generate cross-view images from desired vehicle pairs.

Aihua et al. [89] proposed a framework that primarily comprises view transform and vehicle re-id model. The view transform model comprises of GAN to generate vehicle images in different views to overcome the viewpoint related issue. The vehicle re-id model consists of one backbone, three subnetworks, and one embedding network. The overall framework is illustrated in Figure 15.

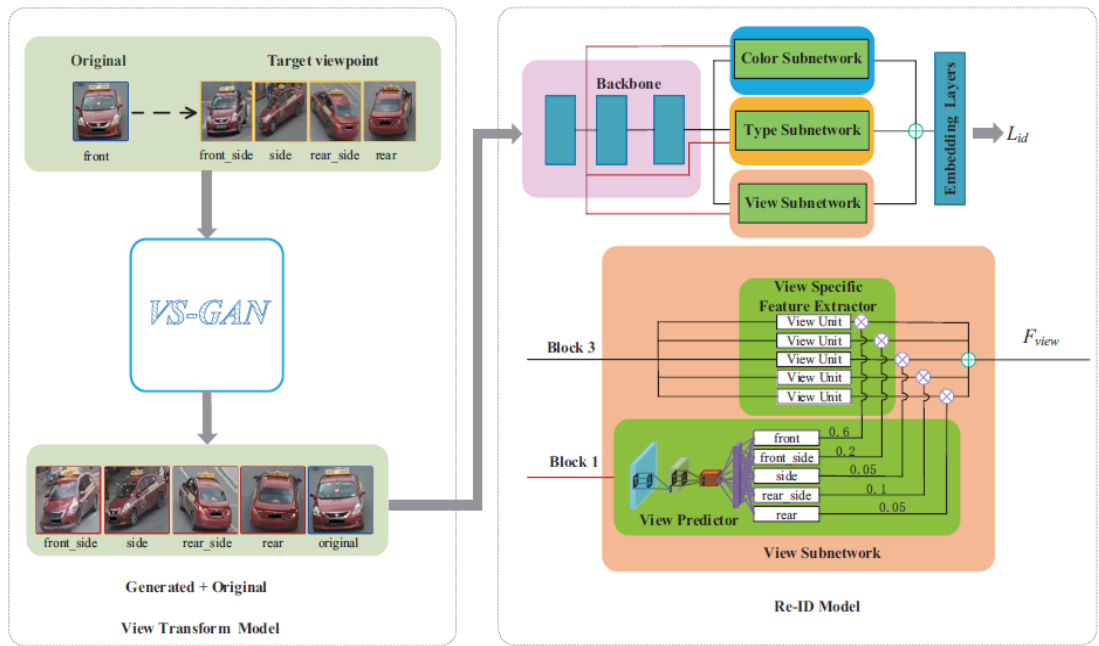

Figure 15. Overview of deep feature representations guided by the meaningful attributes [89].

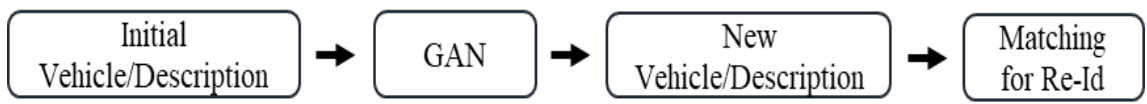

Figure 16. Vehicle re-id system based on GAN diagram.

## 3.7 Attention Mechanism

The neural networks at some extent imitate human brain actions in simple way. Attention Mechanism is also an effort to develop a technique that concentrate on selective thing/actions that are relevant to task and neglecting the others in neural networks. Currently, researchers are trying hard to design an efficient attention-based neural network for vision-related applications. Such as image classification [90], fine-grained image recognition [91], action recognition [92], and re-id [93]. The commonly followed strategy in these approaches is integrating a hard part selection subnet work or soft mask branch into the deep networks. Such as Zhao et al. [94]

studied the part-localization CNN for predicting salient parts and features of these parts  exploit for person re-id. Wang et al. [90] utilizes residual learning technique [51] to  develop the Residual Attention unit for soft mask learning and gained significant image classification results. Though, only the soft pixel-level attention has very small participation in the performance of vehicle re-id task. It gives only global information like vehicle logo, annual inspection stickers, and personalized decorations. So they presented joint learning framework for vehicle re-id in which both soft and hard level attentions are utilized Furthermore, Guo et al. [95] proposed a model with one trunk and two salient part branches for hard part level attention. Trunk branches extracts the global features of vehicle and salient branches extracts the features from vehicle head parts and windscreen. For soft pixel level attention residual attention modules are inserted into trunk and salient branches. Lastly, global and salient part features of vehicle are put to gather for effective feature representation with the supervision of multi-grain ranking loss for vehicle re-id task. Moreover, complete framework is shown in Figure 17.

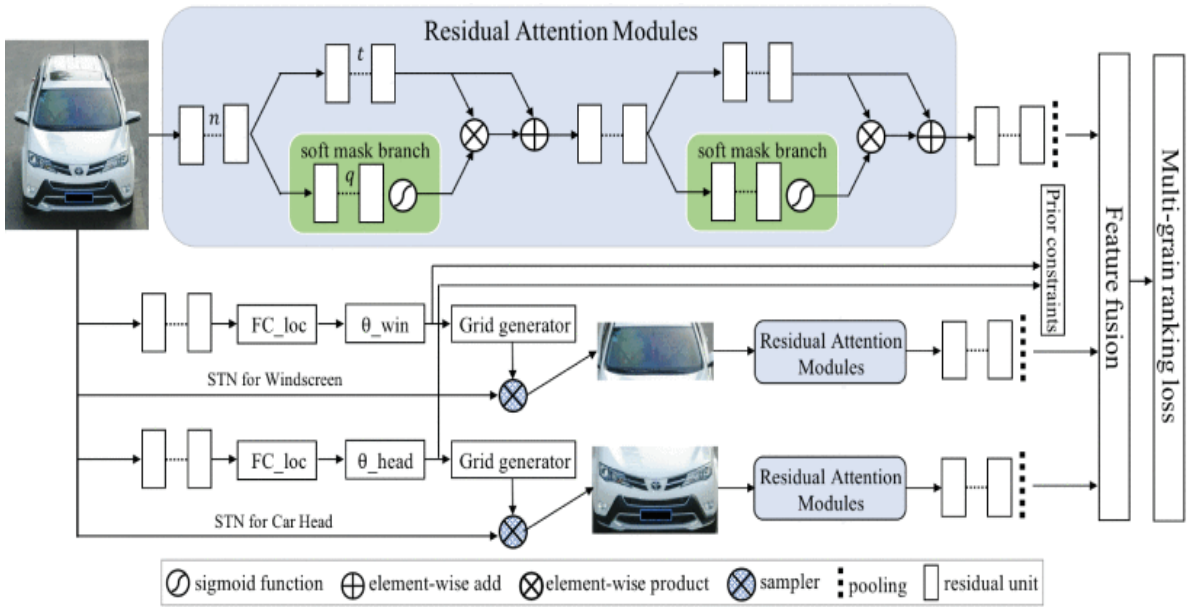

Figure 17. An overview of TAMR structure [95].

Table 2 Comparison of different attention mechanism based approaches

| Method and Reference | mAP% | Rank-1 (%) | Rank-5 (%) |
|:---:|:---:|:---:|:---:|
| RNN-HA [96] | 56.80 | 74.79 | 87.31 |
| SCAN [97] | 49.87 | 82.24 | 90.76 |
| AAVER [98] | 61.18 | 88.97 | 94.70 |
| PGAN [99] | 79.30 | 96.50 | 98.30 |

### 3.8 License Plate-based Vehicle Re-identification

Vehicle re-id using license plate is simply the system's ability to automatically detect, extract, and recognize license plate characters automatically from vehicle image. License plate recognition (LPR) is a conventional method to identify a specific vehicle [100]. An automatic LPR system is mainly divided into two parts, first license plate detection and second, interpreting the vehicle license plate image into numerically readable form. There are many approaches proposed in past for LPR. However, it is still challenging due to some reasons like vehicle image is not captured perfectly, some characters may be occluded, illumination, variation in size of an image, camera distance and zooming. Li and Shen [101] studied a sequence labelling-based approach to recognize the vehicle license plate without character-level segmentation using recurrent neural networks (RNN). The input feature sequence to RNN is extracted using a nine-layer CNN. Super-resolution is also proposed to restore a license plate image to improve performance [102]. Shi et al. [103] designed convolutional recurrent neural network (CRNN) for scene text recognition that incorporates feature extraction, transcription and sequence modeling into a unified framework. Moreover, Figure 18 shows the basic steps of license plate-based vehicle re-id.

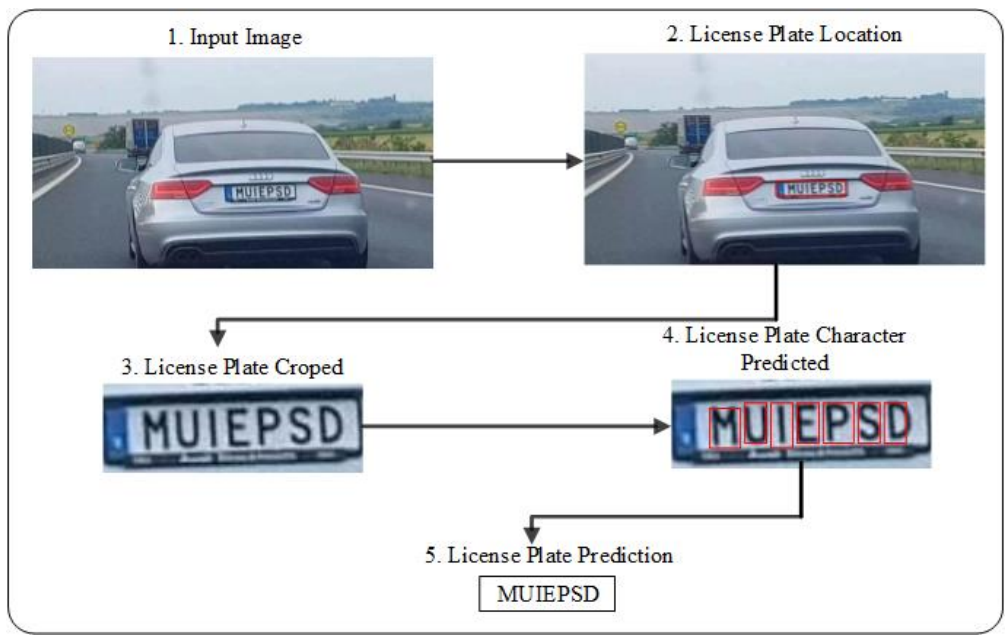

Figure 18. The flow of the license plate based vehicle re-id system.

## 4. Spatio-temporal Cues-based Vehicle Re-identification Approaches

Introducing contextual information in vehicle re-id system can increase the efficiency and reduces irrelevant vehicle gallery images. As compared to person, for vehicle it is necessary to follow traffic rules for instance, practically vehicle follows speed limits, routes, and traffic lanes, so in this scenario vehicle moving in between different cameras at specific time and location helps a lot in vehicle re-id. Spatio-temporal cues are greatly examined for various objects association in surveillance camera network [104]. As in [105] concluded few key findings. Firstly, one specific captured vehicle in one camera cannot appear at more than one location at the same time. Secondly, along the time vehicle is moving continuously based on these finds, authors uses location and time slots to eliminate irrelevant vehicle images from list. Ellis et al. [104] proposed approach that trains the model on temporal and topological transitions of trajectory data and is acquired from surveillance camera network. Loy et al. [106] presented a method for obtaining the spatio-temporal topology of surveillance camera network using multiple camera correlation analysis. Furthermore, time and location information is also exploited for vehicle re-id task [107]. Liu et al. [107] studied a spatio-temporal affinity method for quantifying different pairs of vehicle images. Shen et al. [108] also introduces the spatio-temporal path data for vehicle re-id.

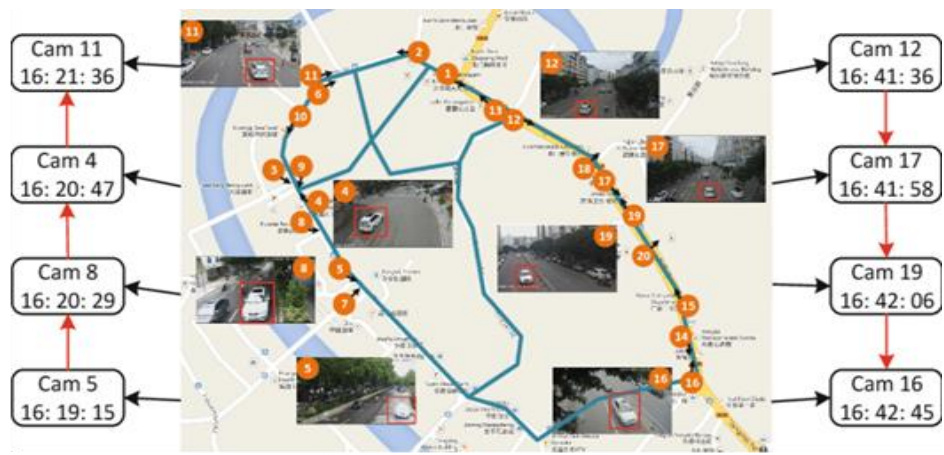

Figure 19. Depicts the spatio-temporal information [107].

## 5. Hybrid Methods Based Vehicle Re-identification

To further enhance the robustness and efficiency of vehicle re-id system researchers have proposed the approaches in which they combined the two or more different techniques, for instance Liu et al. [52] proposed a framework with name PROVID, in this framework author not only consider the visual appearance of vehicle for re-id system, but also exploits the license plate and spatio-temporal cues of vehicle. Jiang et al. [109] studied vehicle re-id algorithm using appearance and contextual information, author examines the multiple attributes during training like vehicle model, color, and vehicle image features individual respectively and sort vehicles on the bases of spatio-temporal cues. Shen et al. [108] designed a two-step architecture, a pair of query vehicle images with contextual information and visual temporal path are produced using MRF chain model, and then the similarity score is generated.

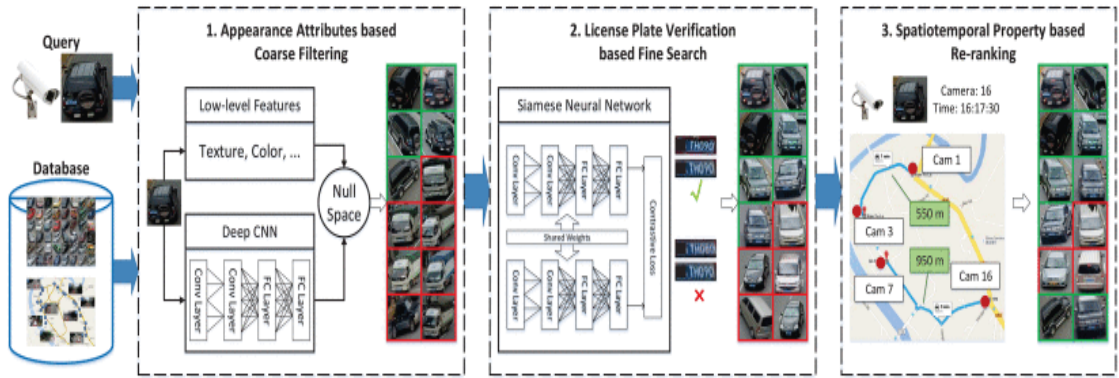

Fig. 20. The architecture of the PROVID framework [52].

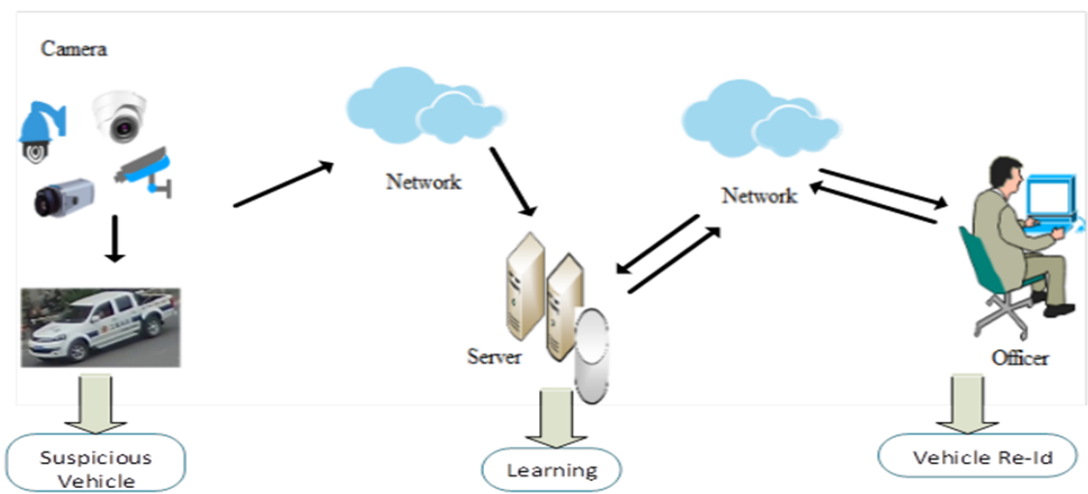

Figure 21. Detailed view of vehicle re-id system.

## 6. Vehicle Re-identification Benchmark Datasets

Datasets are the key components to measure the performance of vehicle re-id system and should reflect the practical surveillance camera data. We cannot avoid some factors like occlusion, background clutter, change in illumination etc. to evaluate the approach [110]. However, multiple benchmark datasets are available, some well-known like VeRi-776, VehicleID, etc. that are prepared by the research community to evaluate vehicle re-id techniques. Table 3 and Figure 30 lists the commonly used vehicle re-id dataset with attributes. Furthermore, a brief description of the most popular datasets is as follows:

***VeRi-776:*** [53] VeRi-776 is a publically available vehicle re-id dataset, and often adopted by the computer vision researcher community. Dataset images are gathered in real scenario using surveillance cameras, and the total images in dataset are 50,000 of 776 different vehicles. Each captured vehicle images have 2 to 18 viewpoints with different resolution, occlusion, and illumination. Furthermore, spatio-temporal relations and license plate are annotated for all vehicles. To make dataset more robust, images are labelled with color, type, and vehicle model. In Figure 22 various types of vehicles from VeRi dataset are shown.

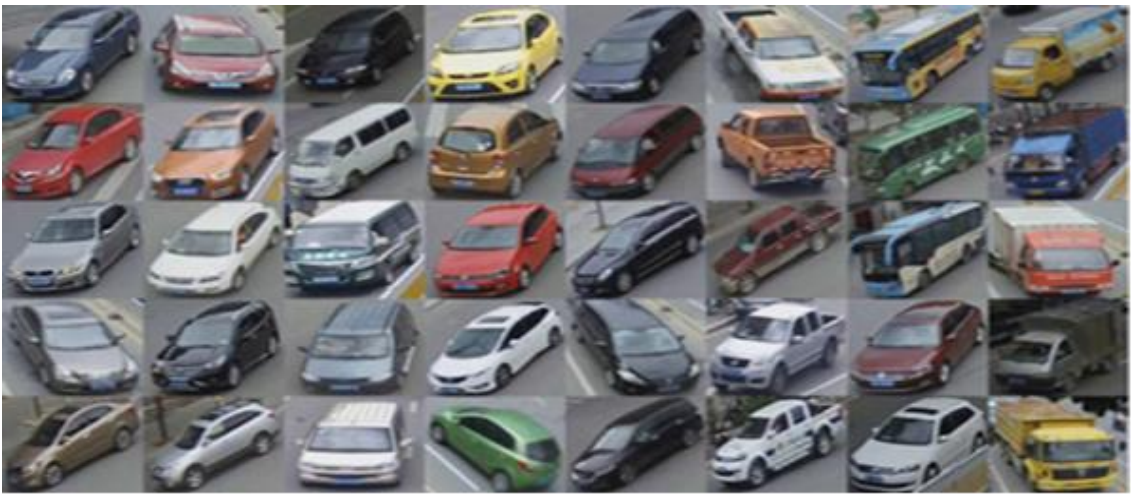

Figure 22. Depicts the sample images of VeRi-776 dataset.

***PKU VehicleID:*** [111] VehicleID dataset is developed by Peking University with the funding of the Chinese national natural science foundation and national basic research program of China in the national engineering laboratory for video technology (NELVT). The vehicle dataset consists of 221763 total images of 26267 vehicles, and all the images are captured during daytime in a small town of China with multiple surveillance cameras with 10319 vehicles model information i.e "Audi A6L", "MINI-cooper" and "BMW 1 Series" are labeled manually.

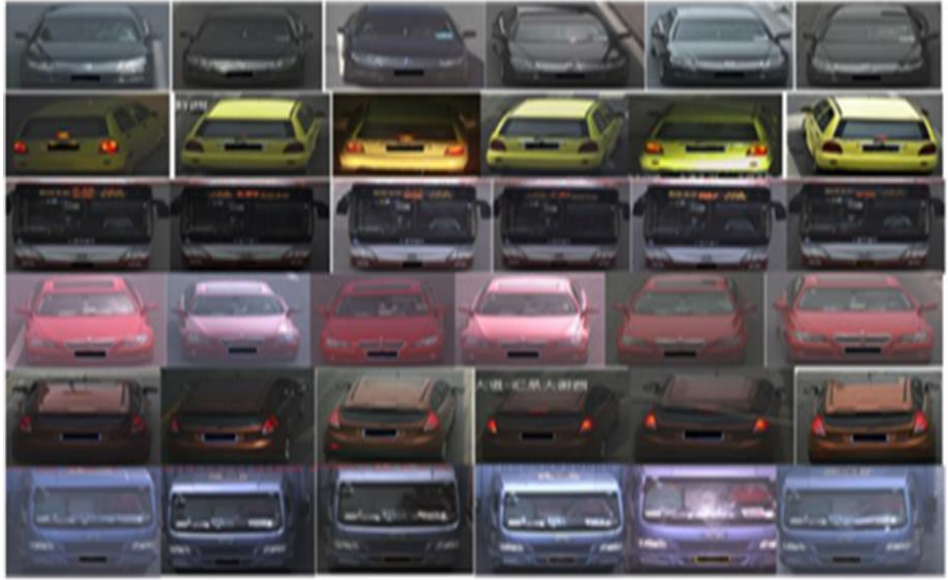

Figure 23. Depicts the sample images of PKU VehicleID dataset.

***Vehicle-1M:*** [112] Vehicle-1M dataset is developed by the University of Chinese Academy of Sciences in the National laboratory of pattern recognition, Institute of Automation. This benchmark dataset contains 55527 vehicles with 400 different vehicle models, and the total captured images are 936051. Surveillance cameras capture all the images in China's town at day and night time and consist of a vehicle's rear and head view. Moreover, each image in this dataset is labeled with a model, make, and vehicle year. Images from Vehicle-1M are shown in Figure 24.

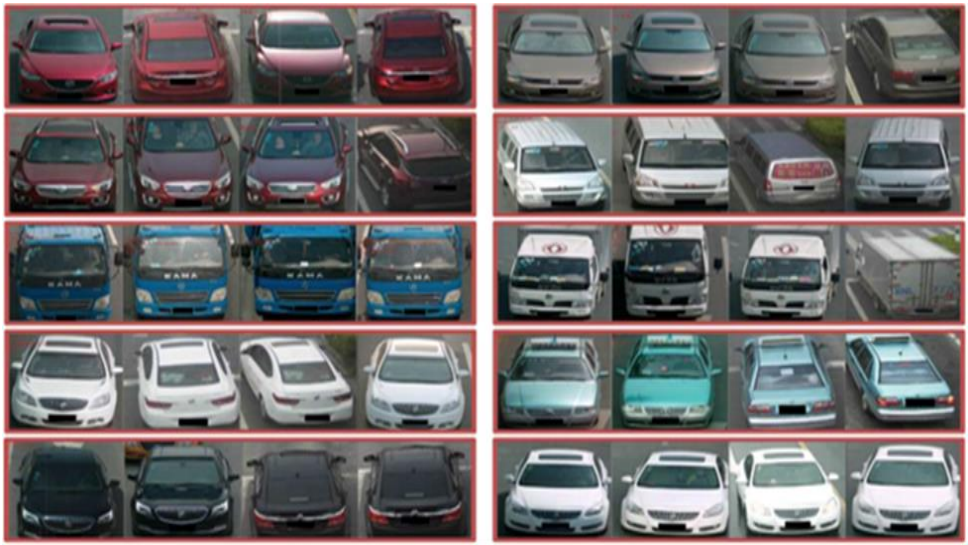

Figure 24. Depicts the sample images of vehicle-1M dataset.

***BoxCars21k:*** [44] BoxCar116k dataset is developed using 37 surveillance cameras, and this dataset consists of total images 116,286 of 27,496 vehicles. For the preparation of dataset, 45 brands of the vehicle are used. Moreover, captured images of the vehicle in the dataset are in an arbitrary viewpoint, i.e., side, back, front, and roof. All vehicle images in the dataset are annotated with 3D bounding box, model make, and type. However, some sample images are shown in Figure 25.

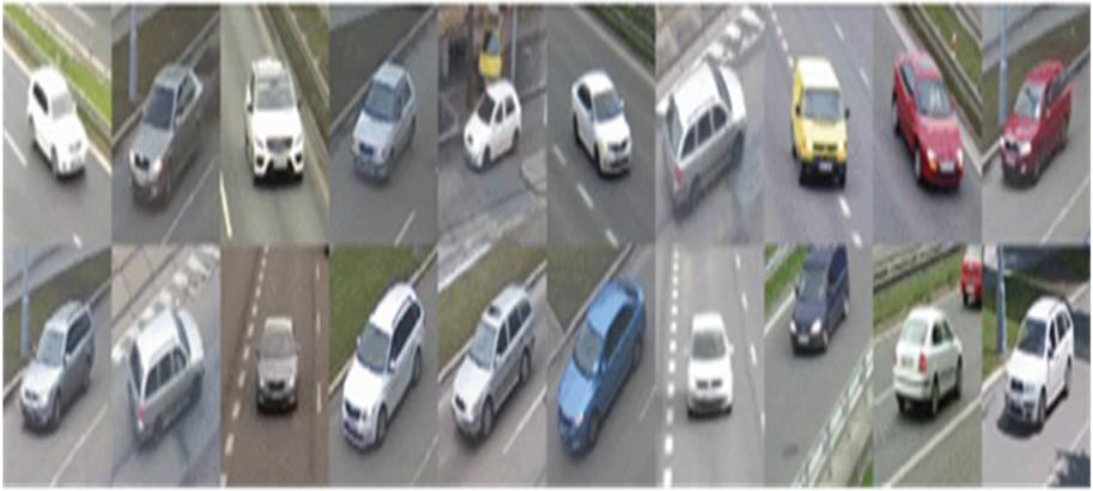

Figure 25. Depicts the sample images of BoxCar21k dataset.

***VehicleReId:*** [64] VehicleReId dataset provides 47123 vehicle images and all these images are extracted from five different video shots by using two surveillance cameras, out of total images 24530 vehicle image pairs are human annotated.

***CompCars:*** [113] CompCars dataset consists of two types of image nature 1)Web-nature images 2) Surveillance-nature images. There are total of 136,726 web-nature images in which there are 163 car makers with 1,716 car models. However, in surveillance-nature, the total car images are 50,000 that are captured from the front view. Samples of CompCars dataset are shown in Figure 26.

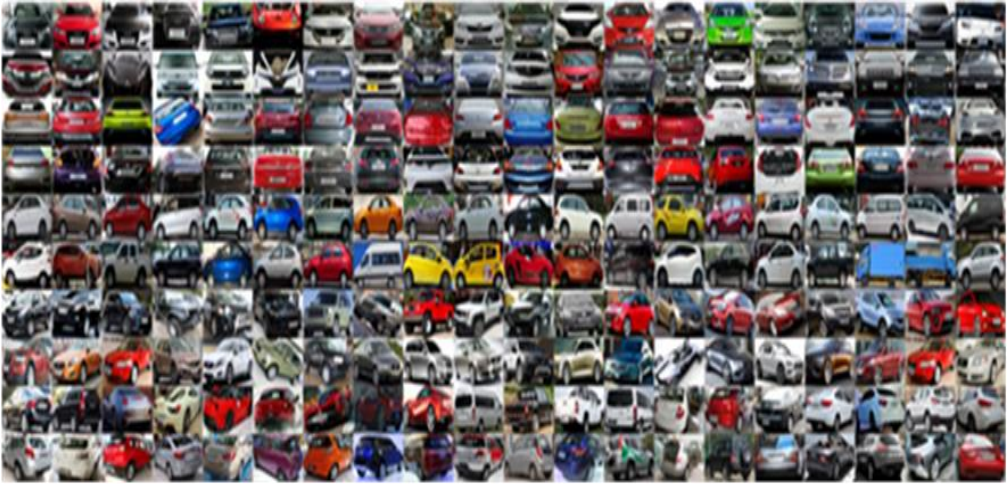

Figure 26. Depicts the sample images of CompCars dataset.

***VRIC:*** [114] VRIC contains 5,622 vehicles with 60,430 total images with different traffic road surveillance cameras and images captured at day and night. Images with different angles, viewpoints, occlusions and illuminations from VRIC dataset are depicted in Figure 27.

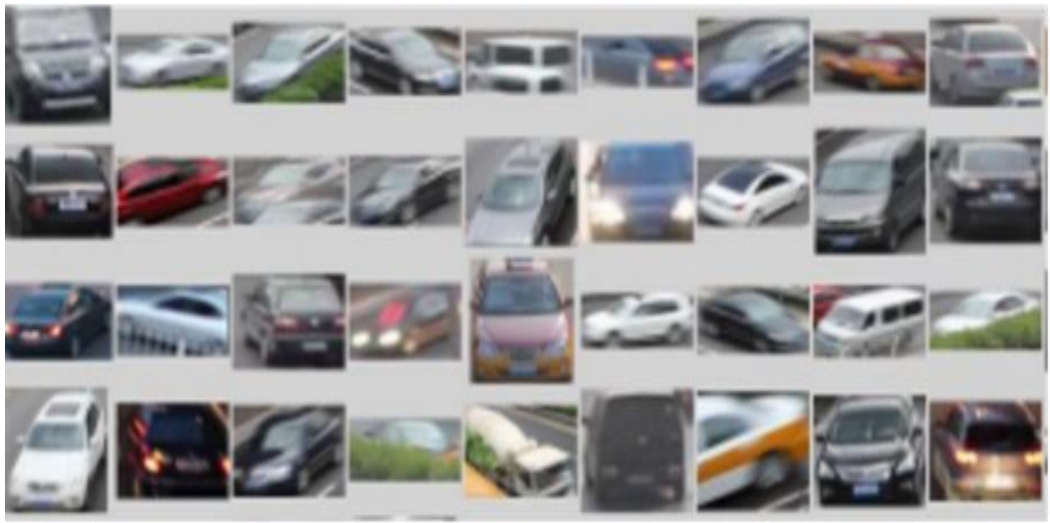

Figure 27. Depicts the sample images of VRIC dataset.

***VRID:*** [115] This dataset contains total 10,000 images and specially developed for vehicle re-id with 326 surveillance cameras the VRID images were captured from 7 am to 5 pm for one week. In the development of the dataset there are 1000 vehicles used with 10 commonly used vehicle models, and at least 10 times each vehicle is captured over a camera network in Guangdong city, China. Surveillance cameras have been fixed in a practical environment with arbitrary directions and angles; therefore, dataset images have various resolutions and poses distributed from 400×424 to 990×1134 pixels.

***VERI-Wild:*** [116] Collects a large-scale vehicle re-id dataset in the unconstrained environment. For dataset development, an existing large CCTV system is utilized. It consists of 174 cameras across, recorded till one month (30×24h). The CCTV cameras are spread over a large city consists of 200 km2. The dataset includes 12 million vehicle raw images, and 11 volunteers cleaned the dataset for one month. After data cleaning and annotation, 416,314 vehicle images of 40,671 identities are collected. VERI-Wild dataset images with viewpoint changes, illumination variations, occlusion, and background variations are presented in Figure 28, and statistics are shown in Figure 29.

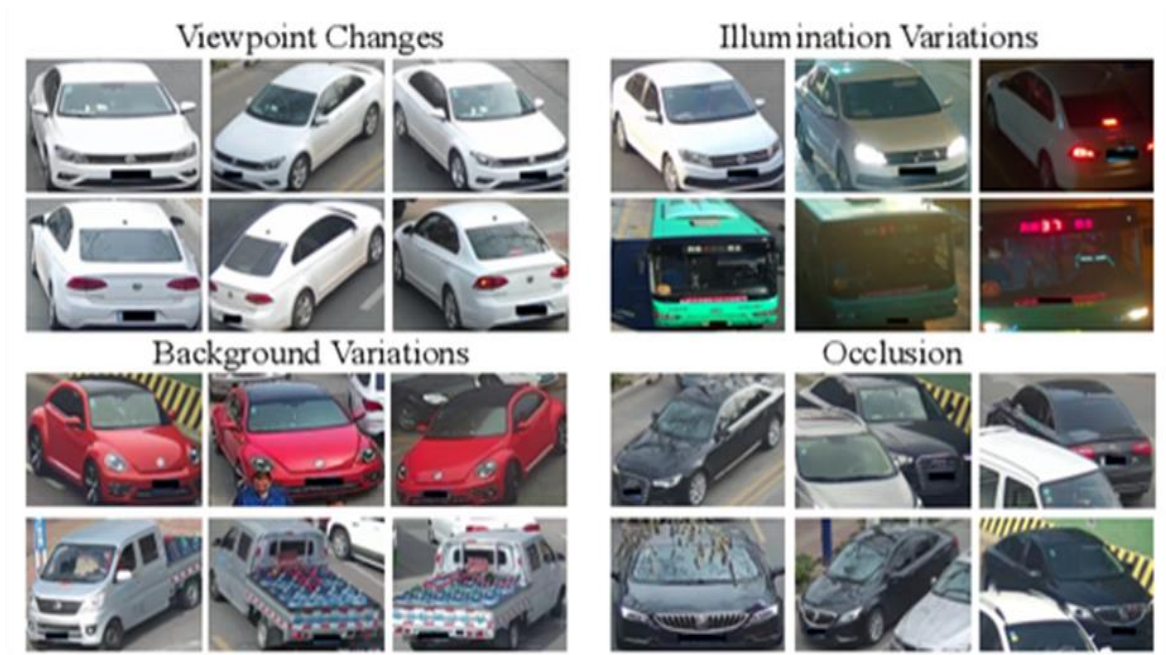

Figure 28. Depicts the sample images of VERI-Wild dataset.

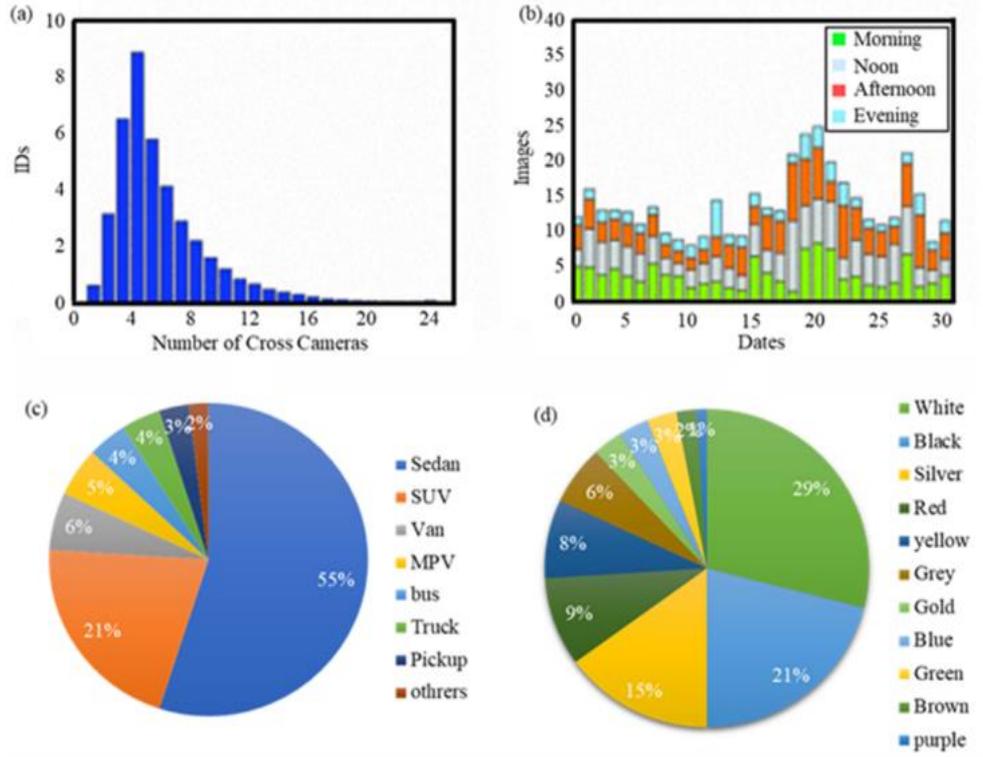

Figure 29.Illustrates the characteristics of VERI-Wild dataset. (a) The number of identities across multiple surveillance cameras; (b) Total number of IDs captured in different slots of each day; (c) Division of vehicle types; d) Division of vehicle colors.

Table 3. Characteristics of publicly available datasets.

| S. No | Dataset | Year | Total no. of images | No. of vehicle models | No. of vehicles | No. of viewpoints | No. of Cameras |
|-------|---------|------|---------------------|-----------------------|-----------------|-------------------|----------------|
| 1 | VeRi-776 [53] | 2016 | 50,000 | 10 | 7,76 | 6 | 18 |
| 2 | PKU VehicleID [111] | 2016 | 22,1763 | 250 | 26,267 | 2 | 12 |
| 3 | Vehicle-1M [112] | 2018 | 936,051 | 400 | 55,527 | …… | …… |
| 4 | BoxCars21k [44] | 2016 | 63,750 | 148 | 21,250 | 4 | …… |
| 5 | VehicleReId [64] | 2016 | 47,123 | …… | 1,232 | …… | …… |
| 6 | CompCars [113] | 2015 | 136,726 | 1,716 | …… | 5 | …… |
| 7 | VRIC [114] | 2018 | 60,430 | …… | 5,622 | …… | 60 |
| 8 | VRID [115] | 2017 | 10,000 | 10 | 1000 | …… | 326 |
| 9 | VERIWild [116] | 2019 | 416,314 | …… | 40,671 | Unconstrained | 174 |

Table 3 sum up the datasets discussed above in terms of year, total number of images, total number of vehicle models, total number of identity vehicles, number of viewpoints, and number of cameras.

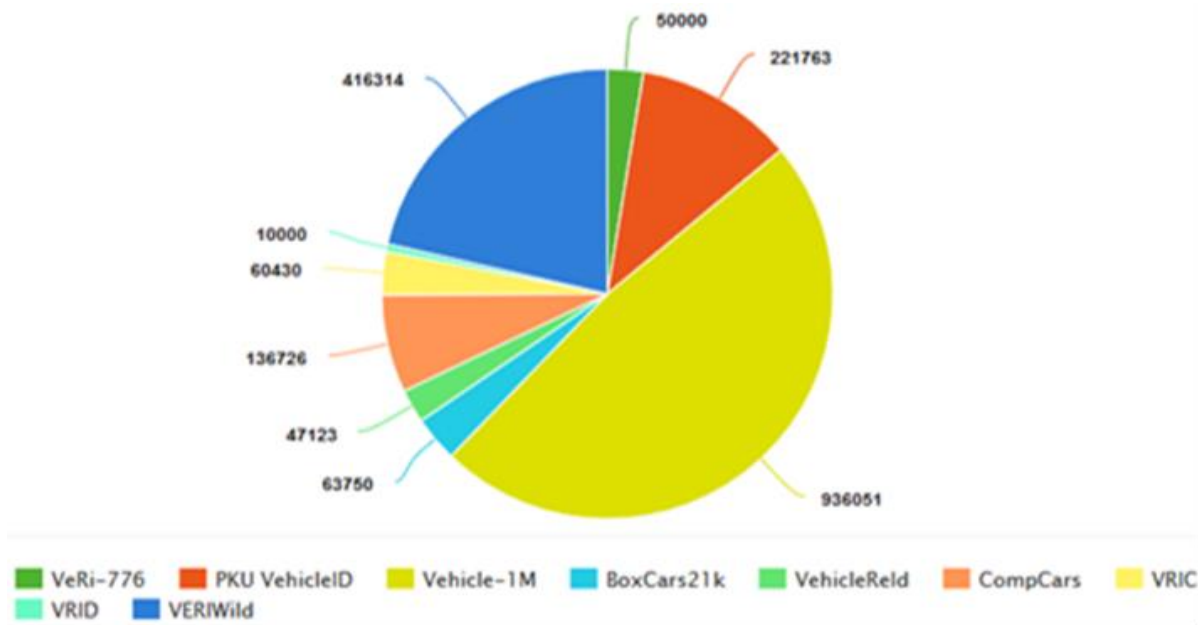

Figure 30. Depicts the number of total images per vehicle re-id dataset.

## 7. Challenges Regarding Vehicle Re-identification

The vehicle re-id is among an essential and challenging task, and it is defined as, either any specific vehicle captured in one camera has already appeared over multiple camera network or not. With the increasing need for automated video analysis, the vehicle re-id receives increasing attention these days in the computer vision research community. Therefore, some key factor and their effects on performance are explained following.

i.   Insufficient data: For vehicle re-id systems each single image should match with gallery images, so it is very hard to get sufficient data for good model learning of each intra-class variability. However, it is also major challenge that dataset should reflect the real-world surveillance, currently, most of the datasets available are consists of non-overlapping views with a limited number of cameras; as a result, datasets have few viewpoints with unchanged regulation, and most of the publically available datasets are consists of limited instances and classes that influence the performances.

ii.  Inter-class similarity: This problem arises due to different automobile manufacturing companies have a similar visual appearance, as a result, two different make, model, and type of vehicles looks similar from rear or front side, as shown in Figure 31. [117][118][119][120].

iii. Intra-class variability: due to the unconstrained environment and viewpoint, the same vehicle looks different over different geographical locations of the surveillance camera network [119][120][14], as depicted in Figure 31.

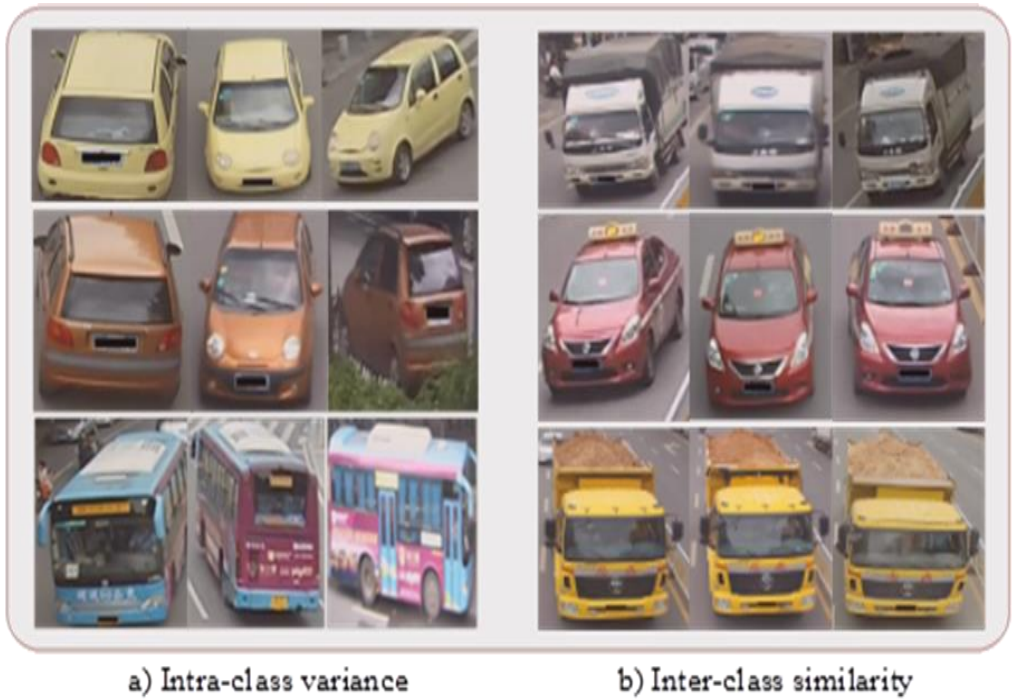

a) Intra-class variance        b) Inter-class similarity

Figure 31. Demonstration of two main challenges in vehicle re-id.

iv. Pose and viewpoint variations: Due to the camera calibration, viewing angle and location on the roadside, captured vehicle image appearance varies, and the same vehicle looks different and different looks same [93][80] . A learned model on the rear pose of a vehicle will probably fail to detect a vehicle's front, side pose. Furthermore, the effect of viewpoint change on vehicle is shown in Figure 32.

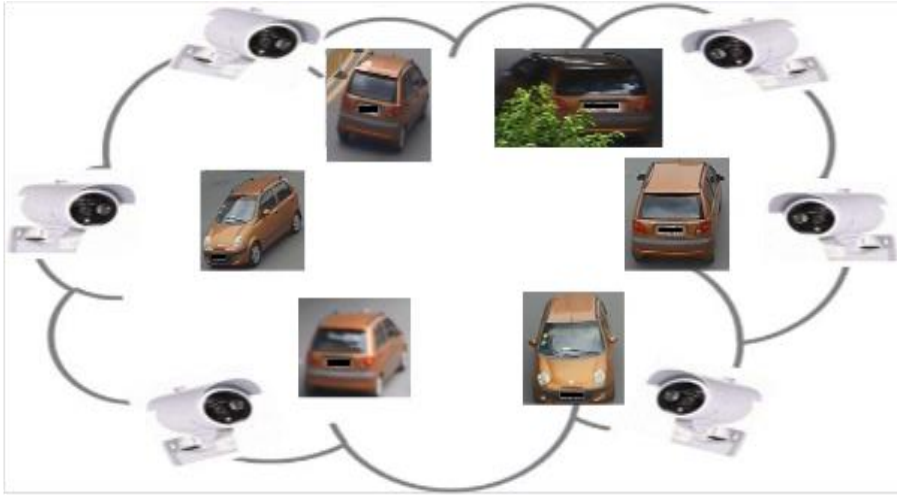

Figure 32. Images of the same vehicle taken from different cameras to illustrate the appearance changes.

v.    Partial occlusions: If some part of an input vehicle is hidden by any object or vehicle in congestion as result, some key discriminative parts are not visible and the matching fails probably. Moreover, due to these features generated by an occluded vehicle image is corrupted [109][121].

vi.    Illumination changes: Vehicle captured images illumination varies surveillance camera to surveillance camera and surveillance camera scenes [14] and also illumination changes on the same surveillance camera due to different time slots like day and night. The same vehicle observed in different lighting conditions can have a color difference on the appearance because of the unconstrained environment [122]. Vehicles lights also have an effect on image illumination, so vehicle appearance changes at different a period of time and multiple camera network [93][109][123].

vii.    Resolutions variation: Changes in resolution in pair of same vehicle occurs because of camera calibration, and another factor is various old surveillance cameras with different heights are fixed on the roadside that give a different-resolution,

viii.    Deformation: Due to load or accident, vehicle shape, and body changes.

ix.    Background clutter: This problem occurs in vehicle re-id when the vehicle's color and image background is the same.

x.    Changes in color response: The color attribute is one of the key parameters in vehicle re-id, but surveillance cameras color response changes because of camera settings features [122].

xi.    Lighting effects: Specular reflection and shadows of the vehicle body generate the noise in vehicle image feature descriptor. If vehicle shadow is larger, there are more chance of inconsistency and noise in feature descriptor. As compared to the practical environment in a controlled environment, the lights and specular reflections can be controlled; but practically, we cannot control shadows, and it is one of the major problems in extracting information from the vehicle image

xii. Long-term re-id: If the same vehicle is captured after a long time or captured at different locations, then ther is a high possibility that the vehicle looks different shape wise due to extra carry load/object.

xiii. Cross dataset vehicle re-id: In vehicle re-id systems training and testing of model is performed on same dataset, but it is practically infeasible, due to significant difference between training and testing data and model may not generalize well.

xiv. Insufficient temporal data: Due to the absence of unconstrained environmental information in datasets, it is impossible to exploit temporal data. However, temporal information can play an important role in the performance of vehicle re-id system.

xv. Vehicle re-id system scalability: Scalability means the system can adapt to varying factors while maintaining the performance, such as storing large gallery sizes that are constantly increasing and computational devices that efficiently analyze data.

xvi. Real-time processing: Practical applications require real-time video processing, and the time constraint is the main challenge in vehicle re-id systems.

xvii. Data labeling: This is a common difficulty in the computer vision field. Training a good model robust to all variations in a supervised way couldn't be done without a sufficient amount of annotated data. For a large camera network, manually collecting and annotating the amount of data from each surveillance camera is expensive.

xviii. A small number of images per identity for training: Since one vehicle may appear very limited times in a camera network, it's difficult to collect much data of one single vehicle. Thus, usually data is insufficient to learn a good model of each specific vehicle's intra-class variability.

xix. A large number of candidates in gallery set: A camera network may cover a large public space, like a parking lot. Thus there can be a huge amount of candidate for a given re-id query, and the number of candidate increase over time. The computation for matching with a large gallery set becomes expensive.

xx. Camera setting: Due to different camera settings and features, the same vehicle image captured by different cameras shows color dissimilarities. There may also be some geometric differences. For example, the shape of a vehicle may be observed with varying aspect ratios.

xxi.    Computation: All the proposed methods are based on deep learning. The computation for the training step with back propagation is more expensive than classical methods. In most cases, a powerful GPU is advisable for training, and more computation and memory resources are thus necessary. In applications with real-time constraints and without GPU, a very deep network may not be suitable for inference.

## 8. Evaluation Metrics

In the re-id task, the target object's images are mostly aligned and cropped. However, the vehicle re-id task is same as the instance retrieval. Given the input image, the candidates with a similar input image in the gallery set are required to be placed in the top positions within a ranking list. To measure the performance of vehicle re-id approaches, the cumulative matching characteristics (CMC) curve is commonly used by researchers. CMC curve provides the probability that an input image identity appears in a different-sized gallery set. The cumulative number of correctly matched inputs is demonstrated based on the rank list in which inputs are re-identified. Where the number of correctly re-identified input images in rank 1 is q(i), the CMC value for rank i can be defined as:

$$CMC(i) = \sum_{r=1}^{i} q(r) \tag{5}$$

where $r$ represents the rank index. CMC curve not only computes the rank-1 but also places the correctly matched images top ranks. Therefore, the CMC curve is a suitable option to describe the vehicle re-id performance of different approaches. Besides, CMC curves, if multiple ground truths for each query image in the gallery set are available, mean average precision (mAP) is used to measure the overall performance for vehicle re-id system. For the given query image, the average precision (AP) can be defined as:

$$AP = \frac{\sum_{k=1}^{n} P(k) \times G(k)}{Ngt} \tag{6}$$

where n is the number of test tracks and Ngt represents the number of ground truths. P(k) shows the precision at cut-off k in the ranking lists. G(k) equals 1 if the k-th match is true, otherwise 0. The mAP measures the overall performance of vehicle re-id system. Therefore, the mAP can be defined as:

$$mAP = \frac{\sum_{q=1}^{Q} AP(q)}{Q} \tag{7}$$

where $Q$ denotes the number of queries.

Another way by which vehicle re-id techniques performance can be evaluated is the confusion matrix. A confusion matrix consists of various columns and rows; it depends on the number of classes. It's diagonal represents the recognizing accuracy or true classification and off-diagonal express the misclassification.

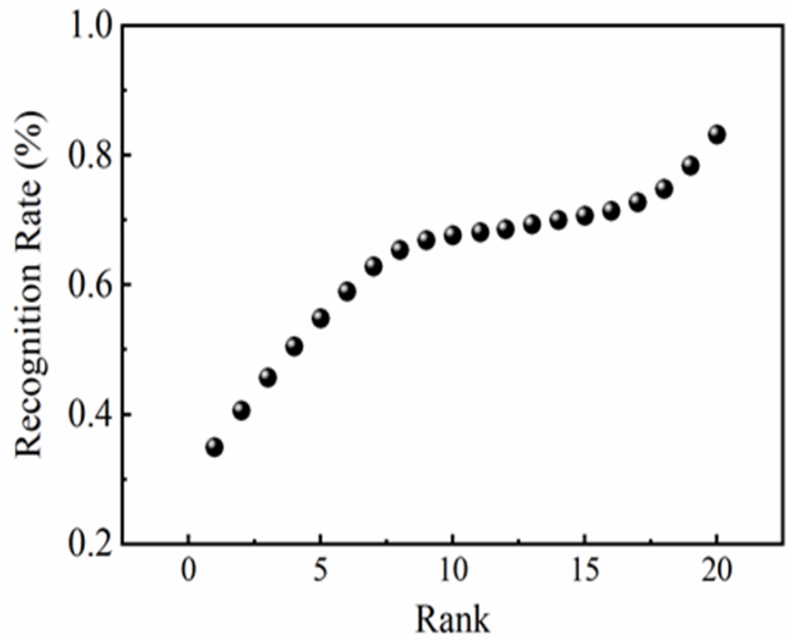

Figure 33. Cumulative matching characteristics (CMC) curve.

## 9. Performance Comparison of Recently Proposed State-of-the-Art Approaches

The first phase in vehicle re-id is to decide whether the given vehicle image exists in the gallery set or not. In other terms, before considering for a similar match, the vehicle re-id system should have the capacity to decide whether the given vehicle probe image is a part of the gallery set or not. This approach is known novelty detection and it needs that vehicle re-id systems to have the ability to discard the miss-matched vehicle images. Usually in vehicle re-id systems, once the gallery set images are ranked in comparison with the given query image, the query image belongs to the gallery set if the similarity distance is higher than an operating threshold. We give a summary of the vehicle re-id mAP of some state-of-the-art methods including CMGN+Pre+Track [124], DF-CVTC [89], PROVID [52], and RAM [125] etc. on VeRi-776 dataset mentioned above. We have chosen VeRi-776 dataset for comparison because it is consisting of varying illumination, more viewpoints, and resolution. In short, this dataset fulfills most of the aspects of real-world camera surveillance data. The statistics about this dataset have been provided in Table 3.

However, Table 4 provides recently proposed state-of-the-art approaches on VeRi-776 dataset. For comparison, we measure the performances of each method in mAP, HIT@1 and

HIT@5 . From Table 4, and Figure 34 we can observe that mAP of different models is increasing during the years 2016 to 2020. As on VeRi-776 dataset from the years 2016 to 2020, the performances of state-of-the-art methods have improved from 12.76% to 85.20%, with an increase of 72.44%. Moreover, Figure 35 shows the CMC of different state-of-the-art approaches on VehicleID dataset with different test size.

Table 4. Performance analysis of some proposed approaches in state-of-art on VeRi-776.

| Reference | Venue | Approach | mAP | HIT-1% | HIT5% |
|---|---|---|---|---|---|
| Year 2020 | | | | | |
| L.Xiangwei et al. [126] | Mobile Networks and Applications | JPFRN | 72.86 | 93.14 | 97.85 |
| Z.Aihua et al. [127] | Neural Computing and Applications | MSA | 62.89 | 92.07 | 96.19 |
| Q.Jingjing et al. [128] | Measurement Science and Technology | SAN | 72.5 | 93.3 | 97.1 |
| W.Honglie et al. [129] | Applied Sciences | LFASM | 61.92 | 90.11 | 92.91 |
| Z.Jianqing et al. [130] | IEEE Internet of Things | JQD$^3$Ns | 61.30 | 89.69 | 95.17 |
| Z.Hui et al. [131] | IEEE ITNEC | AAN+triplet +focal+range (Model-3 | 75.14 | 5.17 | 97.80 |
| O.Daniel et al. [132] | IEEE Access | MidTriNet+UT | ……. | 89.15 | 93.74 |
| L.Sangroket al. [133] | CVPRW | StRDAN (R+S, best) | 76.1 | ……. | ….. |
| J. Zhu et al. [134] | IEEE TITS | QD-DLF | 61.83 | 88.50 | 94.46 |
| L.Xiaobin et.al. [135] | IEEE Trans. on Image Processing | GRF+GGL | 00.61 | 0.89 | 0.95 |
| Year 2019 | | | | | |
| A. A-Acevedo et al. [124] | IEEECVPR | CMGN+Pre+Track | 85.20 | 96.60 | … … |
| F. Wu et al. [136] | Image Communication | SSL+re-ranking | 69.90 | 89.69 | 95.41 |

| S. Ahmed et al. [137] | IEEE ICIP | Mob.VFL-LSTM + Mob.VFL | 59.18 | 88.08 | 94.63 |
|---|---|---|---|---|---|
| G. Rajamanoharan et al. [138] | IEEE CVPR | MTML-OSG | 68.3 | 92.0 | 94.2 |
| P. Khorram et al. [98] | ArXiv | AAVER+ResNet-101 | 61.18 | 88.97 | 94.70 |
| A. Zheng et al. [89] | ArXiv | DF-CVTC | 61.06 | 91.36 | 95.77 |
| Y. Lou et al. [85] | IEEE TIP | Hard-View-EALN | 57.44 | 84.39 | 94.05 |
| J. Hou et al. [139] | Neurocomputing | Baseline + MLL + MLSR | 57.52 | 87.19 | 94.16 |
| B. He et al. [140] | IEEE CVPR | Part-reg. discr feature preserving | 74.3 | 94.3 | 98.7 |
| X. Zhong et al. [141] | ICMM | PGST+visual-SNN | 69.47 | 89.36 | 94.40 |
| R. Kumar et al. [120] | IJCNN | BS | 67.55 | 90.23 | 96.42 |
| Year 2018 | | | | | |
| X. Liu et al. [52] | IEEE Trans. on Multimedia | PROVID | 53.42 | 81.56 | 95.11 |
| Y. Bai et al. [14] | IEEE Trans. on Multimedia | GS-TRE loss W/ mean VGGM | 59.47 | 96.24 | 98.97 |
| J. Zhu et al. [142] | IEEE Access | JFSDL | 53.53 | 82.90 | 91.60 |
| Y. Zhou et al. [81] | IEEE WACV | ABLN-Ft-16 | 24.92 | 60.49 | 77.33 |
| Y. Zhou et al. [80] | IEEE TIP | SCCN-Ft+CLBL-8-Ft | 25.12 | 60.83 | 78.55 |
| N. Jiang et al. [109] | IEEE ICIP | App +Color +Model + Re Ranking | 61.11 | 89.27 | 94.76 |
| J. Zhu et al. [143] | MM Tools and Applications | VRSDNet | 53.45 | 83.49 | 92.55 |
| X. Liu et al. [125] | IEEE IME | RAM | 61.5 | 88.6 | 94.0 |
| D. Xu et al. [122] | ICIMCS | MTCRO | 62.61 | 87.96 | 94.63 |
| D. Sun et al. [144] | Springer ICBICS | ResNet-50 +GoogleNet,+ F.F via CSR | 58.21 | 90.52 | 93.38 |
| S. Teng et al. [97] | Springer PCM | Light_vgg_m+S CAN | 49.87 | 82.24 | 90.76 |
| Y. Zhou et al. [93] | CVPR | VAMI | 50.13 | 77.03 | 90.82 |
| Xiu-Shen et al. [96] | ACCV | RNN-HA (ResNet) | 56.80 | 74.79 | 87.31 |
| Year 2017 | | | | | |
| Y. Zhou et al. [88] | BMVC | XVGAN | 24.65 | 60.20 | 77.03 |
| Y. zhang et al. [56] | IEEE ICME | VGG+C+T | 58.78 | 86.41 | 92.91 |
| Z. Wang et al. [78] | ICCV | OIF+ST | 51.4 | 92.35 | …. |
| Y. Shen et al. [108] | ICCV | Siamese-CNN-Path-LSTM | 58.27 | 83.49 | 90.04 |
| Y. Tang et al. [145] | IEEE ICIP | Combining Network | 33.78 | 60.19 | 77.40 |
| Year 2016 | | | | | |
| X. Liu et al. [53] | IEEE ICME | FACT | 18.75 | 52.21 | 72.88 |
| H. Liu et al. [111] | CVPR | VGG | 12.76 | 44.10 | 62.63 |
| X. Liu et al. [107] | ECCV | FACT + Plate-SNN + STR | 27.77 | 61.44 | 78.78 |
| L. Yang et al. [113] | CVPR | GoogLeNet | 17.89 | 52.32 | 72.17 |

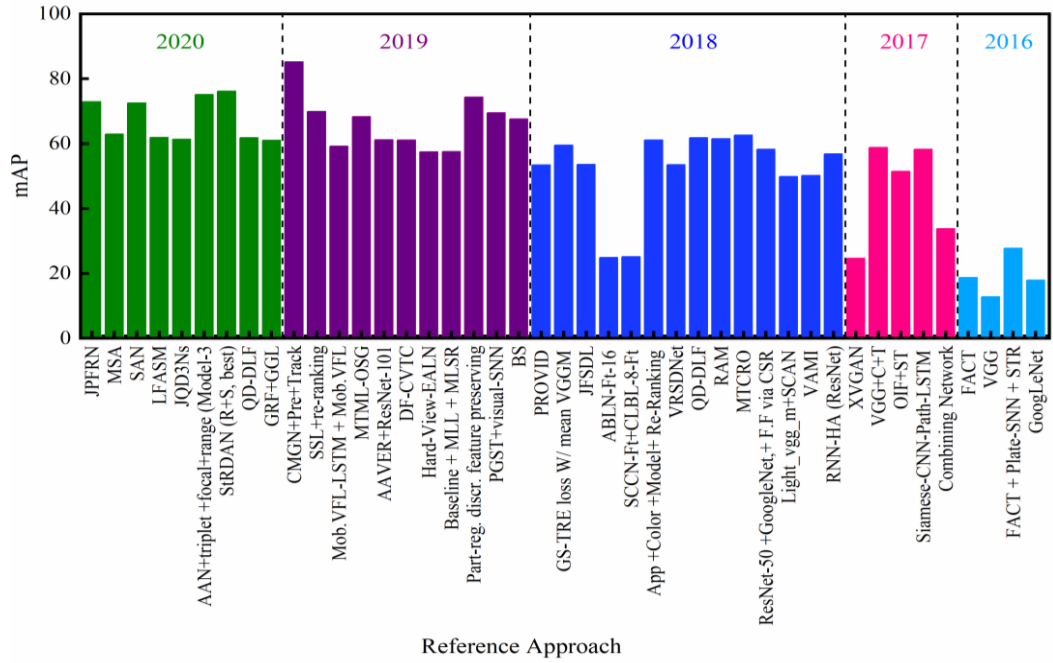

Figure 34. Demonstrates the performance comparison of different state of the art approaches.

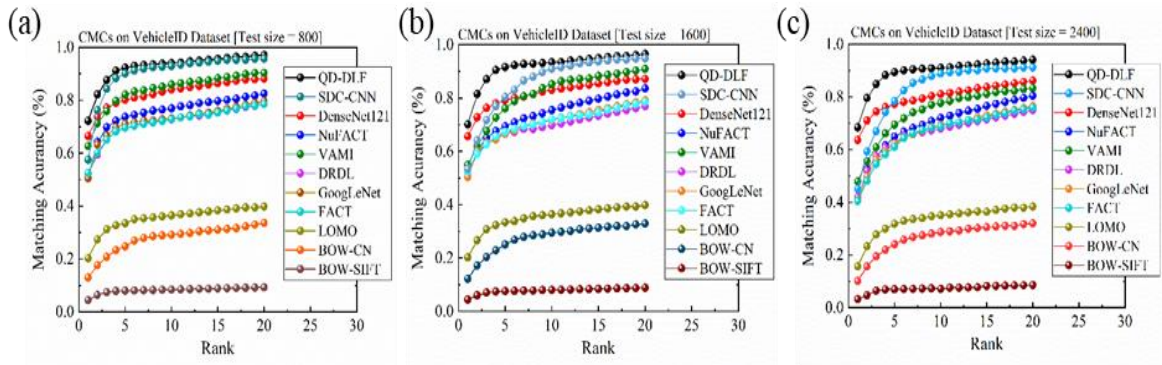

Figure 35. Demonstrates the performance comparison of different state-of-the-art approaches.

## 10. Conclusion & Way Forward

Vehicle re-id is one of the most critical and challenging area in the ITS. Despite high significance, it is not well explored compared to a similar problem that is person re-id. In this review paper, the authors present recent advancements being done for vehicle re-id. Moreover, to draw a detailed picture of study, the authors discuss different vehicle re-id technologies, especially vision-based, including appearance, license plate, spatio-temporal, etc., along with the quantitative and qualitative comparison of different vision-based methods on VeRi-776 and VehicleID datasets. In addition, this review provides comprehensive synopses of publically available benchmark datasets utilized for performance evaluation with a brief description of re-id evaluation techniques. This paper also presents the main challenges as well as applications of camera-based techniques in vehicle re-id.

There are many aspects of vehicle re-id that can be improved. In the future, a reader can explore possibilities to enhance the overall performance of vehicle re-id. Moreover, there is significant potential to extend the approach with some of the following concepts:

CNN works on edges, shapes, and original vehicle features, but the relationship between these features is not considered; hence, the model performance is often unsatisfactory when the vehicle image is rotated or captured with a different rotation. However, a recently capsule network [146] is introduced, which showed improved performance in handling different poses, orientations, and occluded objects.

Secondly, attention-based deep neural network models have gained encouraging results on various challenging tasks, including machine translation [147], caption generation [148], and object recognition [149]. However, attention-based neural network models are still not well investigated for vehicle re-id.

Lastly, due to the development of large-scale real-world data sets, the vehicle re-id system's performance is significantly increased. However, existing datasets offer a specific range of vehicle images with correlated data that causes over-fitting due to over-tuned parameters on specific data. Therefore, the system cannot efficiently generalize other data. A reader can develop large scale real-world surveillance vehicle datasets in an unconstrained environment with multiple views to enhance the training of the state-of-the-art approaches for performance improvement.

Concisely, Vehicle re-id is a demanding and challenging area with massive opportunities for improvement and research. This review paper attempts to provide an overview of the vehicle re-id problem, its challenges, and applications, and, simultaneously, present a way forward. We hope this paper will be valuable for anyone who wants to work in this area.

**Conflicts of Interest**

The authors declare that they have no known competing financial interests or personal relationships that could have appeared to influence the work reported in this paper.

**Funding Statement**

A This paper has been supported The National Key Research and Development Program of China (2017YFC0821505), Funding of Zhongyanggaoxiao ZYGX2018J075 and also supported by Sichuan Science and Technology Program 2019YFS0487.

**Acknowledgments**

I am grateful to my worthy supervisor as well as all the lab mates for their endless support.